\documentclass[11pt]{article}

\usepackage{acl}

\usepackage{times}
\usepackage{latexsym}
\usepackage[ruled,vlined,linesnumbered]{algorithm2e}
\usepackage{amssymb}
\usepackage{amsmath}
\usepackage{mathtools}
\usepackage{booktabs}
\usepackage{multirow}
\usepackage{makecell}
\usepackage{float}
\usepackage{utfsym}
\usepackage{xcolor}
\usepackage{stmaryrd}
\usepackage{colortbl}
\usepackage{tcolorbox}   
\usepackage{enumitem}
\usepackage[normalem]{ulem}
\usepackage{newfloat}
\usepackage{listings}
\usepackage[T1]{fontenc}

\usepackage[utf8]{inputenc}

\usepackage{microtype}

\usepackage{inconsolata}

\usepackage{graphicx}
\usepackage{amssymb}

\title{Deep Interaction: An Efficient Human-AI Interaction Method \\ for Large Reasoning Models}

\author{Hefeng Zhou, Jinxuan Zhang, Jiong Lou, Yuxin Liu, Chaochao Lu, Jingjing Qu$^*$, Jie Li\\
  Shanghai Artificial Intelligence  Laboratory \\
  Shanghai Jiao Tong University\\
  \texttt{qujingjing@pjlab.org.cn} \\}

\begin{document}
\maketitle
\begin{abstract}
The emergence of Chain-of-Thought (CoT) reasoning has significantly enhanced the ability of large language models (LLMs) to tackle complex, multi-step tasks. However, when errors occur, current interaction approaches typically involve re-generating another response that may make mistakes again, or users laboriously flag the faulty step in follow-up turns that may get responses \textit{\textbf{You are right, I made a mistake here}} followed by similar errors recurring. To address this issue, we propose an efficient human intervention mechanism for precisely correcting reasoning errors in LLMs, termed Deep Interaction. Our approach enables direct editing of the original response, allowing erroneous parts to be corrected while preserving accurate reasoning steps. We refine the edited CoT into a distilled prompt, which then steers the LLM along the corrected reasoning path. Experimental results show that our method achieves over a 25\% improvement in correction success rate and reduces token usage by approximately 40\% on STEM tasks reasoning compared to baseline approaches.
\end{abstract}

\section{Introduction}
The advent of reasoning-enhanced large language models (LLMs) have significantly expanded the problem solving capabilities of large language models, enabling them to decompose and address complex tasks through sequential reasoning. Notably, recent iterations such as DeepSeek \cite{liu2024deepseek}, GPT \cite{achiam2023gpt}, and Qwen \cite{team2024qwen2} have demonstrated strong Chain-of-Thought (CoT) \cite{wei2022chain} reasoning abilities, achieving impressive performance on challenging benchmarks. These developments have substantially improved model performance on complex reasoning benchmarks \cite{chen2025towards}. Nevertheless, despite these advances, foundational models still exhibit notable limitations in handling tasks that require deeper understanding or knowledge beyond their training data \cite{wu2024reasoning, mirzadeh2024gsm}. In extensive practical tests, LLMs equipped with CoT succeed on some extremely hard problems. Yet, when they face novel questions that require multiple steps, they still generate answers with clear errors or entirely mismatched reasoning, even if the questions are only at middle-school level. This highlights the need for further improvements in both reasoning capabilities and knowledge grounding.

\begin{figure}    
  \centering
  \includegraphics[width=1\linewidth, keepaspectratio]{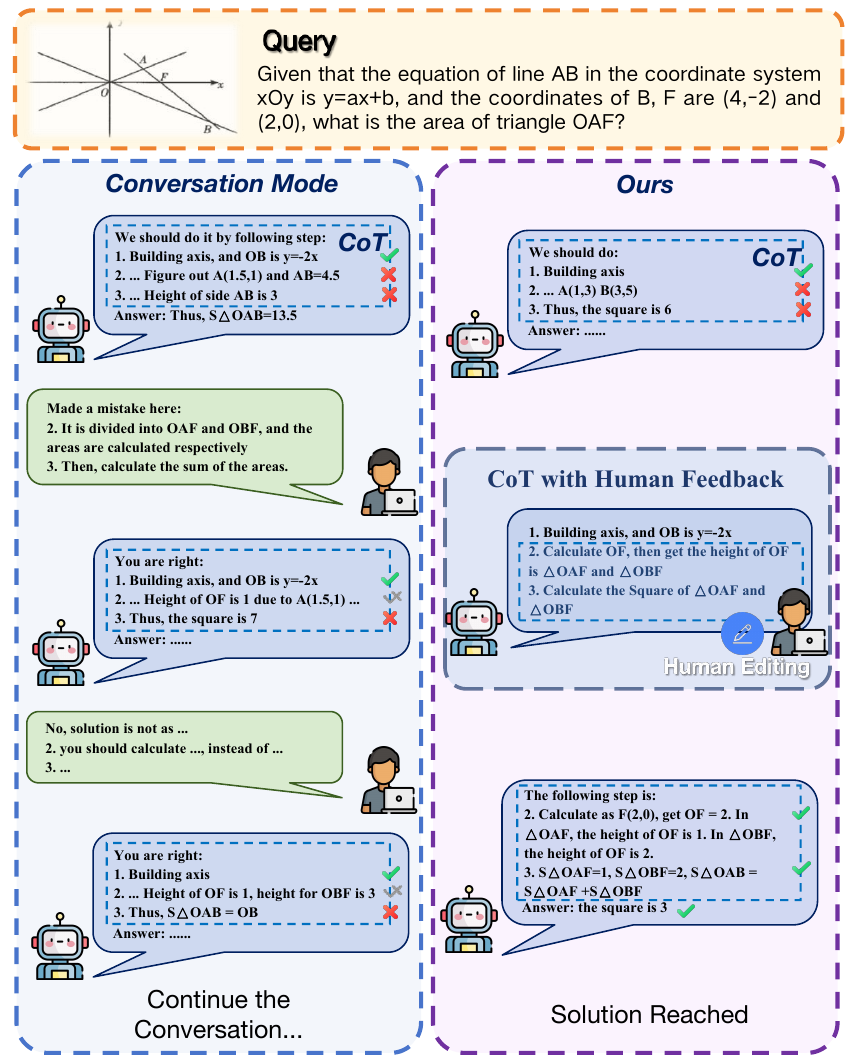}
  \caption{A simple case shows the different pipeline between dialogue-based method and Deep Interaction. Dialogue-based feedback may leave the model on an incorrect reasoning trajectory, while Deep Interaction enables more fine-grained editing of the CoT.}
  \label{fig:pipeline-comparison}
\end{figure}

This study focuses on complex reasoning tasks such as university entrance exams and advanced academic problems \cite{jiawei2025report}. While current LLMs demonstrate strong capabilities in reasoning over three steps or concepts without external tools, they still struggle with longer reasoning chains. Moreover, when corrected or guided with additional knowledge in follow-up interactions, these models often fail to adjust their reasoning paths or follow new instructions effectively \cite{li2024streamingdialogue}. Our analysis of such failures across diverse problem types reveals persistent limitations in robustness and controllability.

\textbf{Instruction Following in Conversational Interaction.} After receiving a response, users typically check its validity. When an error is identified in a specific step within the reasoning chain, they often attempt to point it out in follow-up interactions and guide the LLM toward the correct reasoning path. However, the model may simply acknowledge the mistake — saying something like "You're right, I made a mistake here...", and then repeat the same error in the same step, continuing the reasoning process based on that incorrect foundation \cite{xie2023adaptive,greenblatt2024alignment}. This is a key motivation for our work: users often invest significant effort in identifying and correcting errors, only to receive similar mistakes in return, which can severely undermine their trust in LLMs.

\textbf{Memorization of Training Questions.} As generative models, LLMs predict responses based on learned patterns from training data. Instruction tuning makes them appear more like interactive assistants. However, when faced with questions from their training set, they often \textit{memorize} and regurgitate the original answers, even when minor changes are made to the question \cite{shojaee2025illusion}. This leads to errors when new conditions are introduced. In testing, we found that the model may give correct final answers but with incorrect reasoning steps, or repeat the same answer despite changes in input values.

\textbf{Other Issues. \textit{Performative reasoning}} \cite{ma2025reasoning}. The model may appear to reason step-by-step, but often just stitches together common "solution templates" from training data, lacking true understanding.  It sometimes demonstrates overthinking by producing overly complicated or irrelevant reasoning steps that hinder rather than support problem solving \cite{chen2024not}. \textbf{\textit{Sticky thinking}} \cite{tian2025think}. Early tokens in the input or output can lock the model into a fixed reasoning path, limiting its ability to explore alternative solutions. \textbf{\textit{KV Cache}} \cite{ge2023model} \textbf{\textit{effects.}} Interestingly, on web-based interfaces, even after significant changes to a question’s numbers or semantics, the model may reproduce previous responses verbatim, even in new conversations. We hypothesize that this is influenced by the KV cache mechanism and session-based history retention in these interfaces.

To improve human-LLM collaboration, we propose Deep Interaction, a paradigm that allows users to directly edit reasoning steps like a document revision mode. By leveraging change tracking and semantic refinement, this framework precisely corrects errors and steers model logic without the need for lengthy conversational explanations. Experimental results show over 25\% higher accuracy and a 40\% reduction in token usage on challenging STEM tasks. The method is deployed as the Human-CoT application mode on SafeWork-R1~\cite{safeworkr1}.

\section{Related Work}

\paragraph{Reasoning based Chain of Thought.}
The emergence of large language models (LLMs) represents a significant advancement toward artificial general intelligence. Chain-of-Thought (CoT) methods originates from prompt-based methods \cite{liu2023pre, sahoo2024systematic}, and fundamentally works by following predefined prompting templates to solve problems step by step \cite{brown2020language,kojima2022large}. Since the introduction of CoT concept in 2021, a series of more advanced methods, such as Tree-of-Thought (ToT)\cite{yao2023tree}, Graph-of-Thought (GoT)\cite{besta2024graph}, automated reasoning frameworks \cite{zhang2022automatic}, and Multi-agent CoT (MCoT) \cite{zhang2023multimodal,mitra2024compositional} have been proposed to further enhance the capabilities of LLMs. The release of OpenAI's o1 series marks a key milestone in the adoption of CoT-based reasoning models as the mainstream approach in commercial LLMs services. Currently, model series such as DeepSeek-R1 \cite{guo2025deepseek}, Qwen 2.5 \cite{bai2025qwen2}, and Gemini 2.5 \cite{comanici2025gemini} also exhibit strong reasoning capabilities and demonstrate impressive performance on complex tasks\cite{rein2024gpqa, aleithan2024swe, phan2025humanity}.

Foundation Model of LLMs possess vast knowledge and strong generalization, solving various human tasks. To enhance capabilities, techniques such as Retrieval-augmented Generation (RAG), agents, and CoT have been developed, further improving performance on complex problems.

\paragraph{Correction of Generated Content.}

LLM outputs often suffer from hallucinations \cite{ji2023survey,huang2025survey} and factual inaccuracies \cite{evans2021truthful}, motivating verification and correction methods \cite{manakul2023selfcheckgpt,mishra2024fine}. A line of work studies self-feedback and self-correction: Self-Refine \cite{madaan2023selfrefine} iteratively improves outputs using feedback generated by the same model; RCoT \cite{xue2023rcot} detects inconsistencies by reversing the reasoning chain; and Mind's Mirror \cite{liu2023mind} distills self-evaluation and multi-chain reasoning abilities. Agent-based methods such as ReAct \cite{yao2023react} and Reflexion \cite{shinn2023reflexion} further combine reasoning with actions, observations, or verbal feedback memories. Human-in-the-loop CoT methods expose intermediate reasoning to users for inspection and guidance \cite{cai2023human}. However, prior studies show that LLM self-correction is reliable mainly when errors are easy to detect \cite{huang2023large}, and even strong models often fail to locate their own reasoning mistakes \cite{tyen2023llms,huang2023lvlms}. Therefore, external feedback remains a practical direction for correcting LLM reasoning \cite{kamoi2024can,saha2023branch,wu2024large}. Unlike dialogue-based feedback or automatic self-refinement, our Deep Interaction allows users to directly edit erroneous CoT spans while preserving correct reasoning steps.

\section{Methodology}

\paragraph{Overall Framework.}
Current conversational correction is often inefficient; long dialogue histories can lead models to overlook key information \cite{ConversationalHistory}, while iterative back-and-forth for error fixing remains tedious. From this perspective, we propose a revision paradigm that allows users to directly edit LLM outputs. This approach eliminates the need to describe error locations through additional dialogue, enabling more efficient and targeted corrections.

\begin{figure*}
    \centering
    \includegraphics[width=0.9\linewidth]{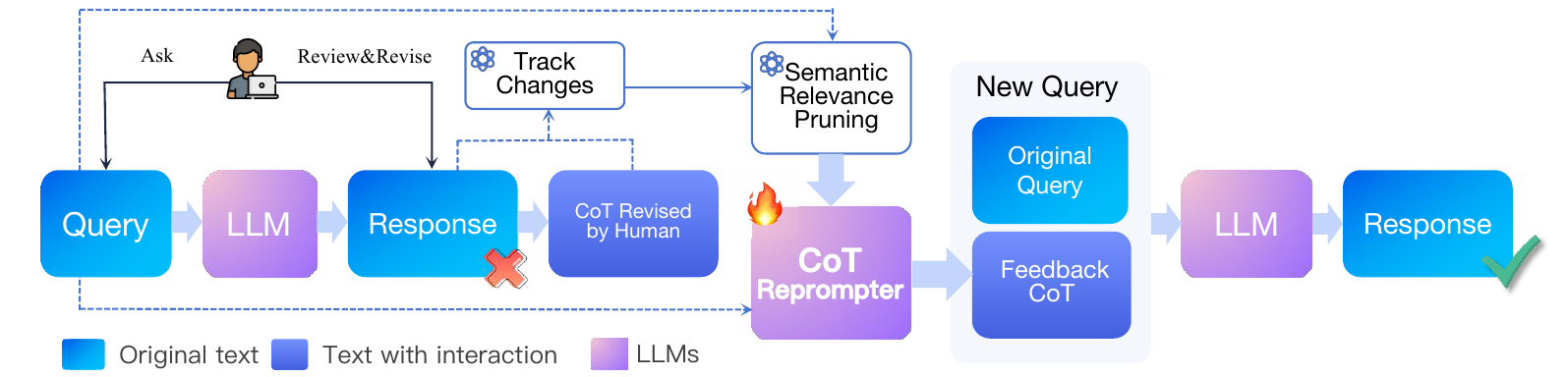}
    \caption{Deep Interaction: An Efficient human intervention framework for human-LLM interaction.}
    \label{fig:enter-label}
    \vspace{-10pt}
\end{figure*}

As illustrated in Figure \ref{fig:enter-label},the process begins with a user query and an initial Chain-of-Thought (CoT) response. If the user identifies a clear error or a deviation from their preferences in the response, they can directly edit the output—similar to using a document revision mode. A track-changes mechanism, based on text differencing, detects the modifications and segments the edited response into parts (deleted, inserted, or retained content). Each segment is then processed in context with the original query before being recombined. Optionally, a CoT reprompter reformulates this content into a structured "Feedback-CoT," which is combined with the original query to guide the LLM in generating a corrected response.

\subsection{Response Mode with Revision}

Human guides reasoning rather than give direct answers—a core principle of our approach. Users correct reasoning steps that contain obvious errors or deviate from their expectations, and may also provide guidance for subsequent inference steps. Building on this idea, consider the r-th round of interaction with a challenging query $\mathcal{Q}_r$ The model's response consists of a CoT reasoning sequence and a final answer. After user editing, the corrected CoT is denoted as $\mathcal{C}_r$. Thus, human intervention yields a refined reasoning path $\mathcal{C}_r'$, while the model retains the ability to perform detailed deduction in later steps.

\begin{equation}
    \mathcal{C}_r' \leftarrow  Human Edit(\mathcal{C}_r).
\end{equation}

Ideally, LLMs should perform step-by-step reasoning, pausing at each stage for confirmation or correction before proceeding (i.e., enabling subsequent reasoning in $\mathcal{C}_r^{sub}$). However, current model architectures do not natively support this mechanism, and such behavior can only be simulated through concatenated step-by-step inference. While we aim for end-to-end problem solving with minimal human intervention, one-shot generation remains the most practical solution today. Future work will explore integrating staged output and human interaction when the model is uncertain.

We aim to build a feedback-driven reasoning mechanism by incorporating human-modified responses into the next round of inference. However, direct input of revised content risks shifting the model’s focus away from the original question $\mathcal{Q}_r$, especially in long conversations. To mitigate this, we assume $\mathcal{Q}_r$ is weakly dependent on prior context and that all necessary knowledge is included in  $\mathcal{C}_r$, or added by user edition. Thus, the new input is formed from the joint representation of $\mathcal{Q}_r$ and $\mathcal{C}_r'$, where $\mathcal{C}_r'$ contains the required prior knowledge.
\begin{equation}
    \mathcal{Q}_{r+1} \gets concat(\mathcal{Q}_r , \mathcal{C}_r').
\end{equation}
Finally, the new response generated, denoted as $\mathcal{C}_{r+1}$ and 
$\mathcal{A}_{r+1}$, represents the updated reasoning path and answer to original $\mathcal{Q}_r$, obtained via human-in-the-loop intervention.

\subsection{Refinement based on Tracking Changes}
LLMs often fail to self-correct effectively because token-by-token generation tends to favor memorization over genuine reasoning. To address this, we introduce human feedback to revise the reasoning path $\mathcal{C}_r$. Our experiments show that flawed $\mathcal{C}_r$ sequences typically contain both correct and incorrect steps, usually following a three-part pattern: an initially correct segment, a middle segment with clear errors, and a later segment whose overall strategy may remain plausible but is contaminated by earlier mistakes. In practice, users mainly edit the erroneous middle segment, while the subsequent content may be deleted, partially revised, or left unchanged due to cognitive effort.

Given the need to trace user modifications to the CoT content within responses, we adapt the Myers Diff algorithm \cite{myers1986nd} as a basic approach for implementing fine-grained text change tracking. The methodological framework for this tracking mechanism is outlined as follows. Give the Original CoT text $\mathcal{C}_r$, and after user edit text $\mathcal{C}_r'$. $\mathcal{T}(\cdot)$ is the tokenization operation. The text editing of the $k$-th segment is denoted as $\Delta_k$, which is represented by starting token index $s_k$, ending token index $e_k$, corresponding operations $o_k$. After computation, we obtain the corresponding edition set $\Delta(\mathcal{C}_r, \mathcal{C}_r')$ as Eq.\ref{eqsets}.

\begin{align}
&\Delta_k = \bigl\langle s_k,\ e_k,\ o_k,\ \text{text}_k \bigr\rangle,\\
&o_k \in \{\,\text{equal},\ \text{delete},\ \text{insert},\ \text{replace}\,\},\\
&\Delta(\mathcal{C}_r,\mathcal{C}_r') =\{\Delta_k\}_{k=1}^{n'} \label{eqsets}.
\end{align}

We use $\mathcal{D}(\mathcal{C}_r,\mathcal{C}_r')=\frac{1}{L}\sum_{k=1}^{n'} (e_k-s_k)\cdot w(o_k)$ to measure text distance, where L is the number of sentences and w is the operation weight (1 for insertion/deletion, 2 for replacement). Note that human edits may still affect subsequent steps. We therefore split the edited CoT into three segments: [pre-edit, edit-section, post-edit], and apply distinct processing strategies to each. Although these operations are lightweight and practical, their rationale and effectiveness are further discussed in the Appendix~\ref{analysis}.

\textbf{Emphasis(.):} To enhance the model's awareness of user edits, modified content should be explicitly emphasized to draw the model's attention. Potential strategies include Markdown formatting, capitalization, or prefix tags such as [IMPORTANT] \cite{carton2021learn}. We adopt double asterisks (**Content**, **) to delimit and highlight edited text as the default emphasis mechanism.

\textbf{Pruning(.):} We use the inherent markdown structure of LLM outputs, splitting text by line breaks or periods. Semantic similarity is computed using sentence embeddings, while approximately 20\% of low-value phrases (e.g., \textit{in this problem}, \textit{to solve this}, \textit{next}, \textit{check}) are filtered out. A theoretical motivation for removing semantically redundant spans while preserving task-relevant meaning is provided in Appendix~\ref{analysis}.

\textbf{Removing(.):} Intermediate step errors often propagate forward, though some cases involve only computational mistakes with correct logic. Users typically delete subsequent content after edits—yet omissions can cause inconsistency. To mitigate this, we retain only section headers (if structured), as they provide reasoning cues; otherwise, we remove the content entirely to prevent confusion from incomplete deletion.

\textbf{Delexicalization(.):} Matching numerical values from training or original outputs can induce memorization errors. We mitigate this by masking numbers to promote semantic reasoning. Appendix~\ref{analysis} further discusses why semantically equivalent but textually perturbed inputs can help avoid over-reliance on fixed surface forms.

Finally, these operations collectively yield the pipeline depicted in Figure~\ref{fig:enter-label}, which is further formalized in Algorithm~\ref{alg:refine1}.

\setcounter{AlgoLine}{0}
\begin{algorithm}[t]
\caption{CoT Edition Pipeline}\label{alg:refine1}
\KwIn{Dialogue history $Dialog_{0..r-1}$,\; User query in r-th round $\mathcal{Q}_r$}
\KwOut{Updated dialogue history $H_r$}
\small
$History \leftarrow Dialog_{0..r-1}$\;


\BlankLine
\tcp{Input query and LLM response}
$(\mathcal{C}_r,\mathcal{A}_r) \leftarrow \text{LLM}(H \oplus \mathcal{Q}_r)$\;

$\mathcal{C}_r' \leftarrow \text{HumanEdit}(\mathcal{C}_r)$\;

[\underline{pre-edit}, \underline{edit-section}, \underline{post-edit}] $\leftarrow \text{TrackChanges}(\mathcal{C}_r, \mathcal{C}_r')$\;

\BlankLine
\tcp{Sentence-level text segmentation}
$\mathbb{T} \leftarrow \mathcal{T}$ ([\underline{pre-edit}, \underline{edit-section}, \underline{post-edit}])\;

\For{$i \leftarrow 1$ \KwTo $len(\mathbb{T})$}{

    \uIf{$t_i$ is pre-edit}{
        $t_i \leftarrow \text{Prune}(t_i)$,\;
        $t_i \leftarrow \text{NumMask}(t_i)$\;
    }
    \uElseIf{$t_i$ is edit-section}{
        $t_i \leftarrow \text{Emphasize}(t_i)$\;
    }
    \uElseIf{$t_i$ is post-edit}{
        $t_i \leftarrow \text{Remove}(t_i)$,\;
        $t_i \leftarrow \text{NumMask}(t_i)$\;
    }
}
\BlankLine
\tcp{Concatenate processed segments}
$\mathcal{C}_r' \leftarrow \text{concat}\left(\bigparallel_{i=1}^{|\mathbb{T}|} t_i\right)$

\BlankLine
$Q_{r+1} \leftarrow concat(Q_r,\mathcal{C}_r')$\;

$(\mathcal{C}_{r+1},\mathcal{A}_{r+1}) \leftarrow \text{LLM}(Q_{r+1})$\;

\BlankLine
\tcp{Output new response, update history later round}

\Return $(\mathcal{C}_{r+1},\mathcal{A}_{r+1})$\;

$H \leftarrow H \oplus (\mathcal{C}_{r+1},\mathcal{A}_{r+1})$\;
\end{algorithm}

\subsection{Semantic-Equivalent Transformations}
As previously mentioned, large language models fundamentally perform autoregressive completion over the input query. Our preliminary experiments reveal that, despite the proposed editing mechanisms, the revised input may still be too similar to the original, leading the model to regenerate previously incorrect continuations (analysis in Appendix~\ref{analysis}). To illustrate: suppose the original response to query $\mathcal{Q}_r$ is structured as [$Step_1\color{green}{\usym{2714}}$, $Step_2\color{red}{\usym{2717}}$, $Step_3\color{red}{\usym{2717}}$, $Answer\color{red}{\usym{2717}}$]. After editing, the new input $\mathcal{Q}_{r+1}$ becomes [$Question$, $Step_1\color{green}{\usym{2714}}$, $Step_2'\color{green}{\usym{2714}}$]. In this case, the model may simply complete the sequence by regenerating the prior erroneous suffix. The new response is [$Step_3\color{red}{\usym{2717}}$, $Step_4\color{red}{\usym{2717}}$, $Answer\color{red}{\usym{2717}}$] and the final answer, due to strong sequential memorization.

Moreover, manually edited content often remains redundant. We therefore introduce a LLM-based feedback trace distillation (CoT Reprompter) that preserves semantics while reducing redundancy and improving clarity. It can be implemented in two ways: \textbf{Few-shot with a extremely large-scale LLM.} We create few-shot triplet samples consisting of the original query, the initial edited CoT from Algorithm \ref{alg:refine1}, and the optimized CoT validated for response correctness. The prompt is \textit{Please transform the following task based on the triplet samples \{...\}: The new Query\{$\mathcal{Q}$\}, the new edited CoT\{$\mathcal{C'}$\}, please output the new optimized CoT.} \textbf{Fine-tune with a lightweight LLM. Similarly.} We construct a triplet dataset automatically generated via the pipeline of Algorithm \ref{alg:refine1}, and fine-tune a lightweight LLM to acquire the reasoning refinement capability. This approach offers faster inference, making the backend processing time imperceptible to users. The method for obtaining the training data is described in Appendix~\ref{finetunedata}.

\section{Experiments}
\subsection{Setup}
\textbf{Models.}
For the problem-solving model, we primarily tested using the latest Qwen-VL-Max via API. Additionally, for comparison, we selected models with similar performance: GPT, Gemini \cite{team2023gemini}, and Claude \cite{claude}. In the CoT Prompter, the model used for the few-shot context learning path is Qwen-VL-72B, while for the fine-tune path, we used Qwen-VL-7B as the base model. Both the experimental environment and model fine-tuning were conducted on cloud servers equipped with 64 core CPUs, 400GB of RAM, and 4 $\times$ A100 GPUs.

\subsection{Datasets}
We adopt a multi-dimensional evaluation approach, using general safety alignment cases, logic questions and simple yet error-prone STEM questions for case studies. For quantitative evaluation, we select ScienceQA \cite{lu2022learn}, Gaokao-MM \cite{zong2024gaokao} and LogicQA~\cite{kwon2025logicqa} as benchmarks, which provide annotated solutions and correct answers. Additionally, we curated 5,000 high-difficulty questions in each of mathematics, physics, chemistry, and biology. Every question is accompanied by a detailed solution. The items span college-entrance to university-level curricula and exceed the complexity of Gaokao-MM. We designate this corpus STEM20K. This dataset will be released later, with a subset provided in the supplementary materials.

\subsection{Evaluation Metrics}
\textbf{Comparison Method.} We compare our method with existing dialogue-based correction approaches. For challenging problems, neither dialogue-based correction nor CoT editing can reliably guarantee correctness in a single round. We therefore adopt a multi-round evaluation protocol to assess whether the system can eventually reach the correct answer through iterative dialogue or edits. We further introduce an expert mode (\textit{-Expert}), in which professional users may perform partial calculations manually and use the LLM for verification. This mode also supports the generation of standardized instructional solutions, while large-scale reasoning is evaluated by an LLM-based judge calibrated against expert annotations.

\textbf{Pass Rate.} For STEM questions, a response is considered passed if the final answer and reasoning are fully correct. A direct pass indicates correctness in the first round. Success achieved through iterative dialogue or editing within the $x$-th round is recorded as pass within xR. During testing, we run two rounds each dataset. If in either of the two rounds, the conversation first completely answers the solution approach and the answer correctly, it is considered as passed.

\textbf{Correction Rate.} Since fully correcting multi-step problems in a single round is difficult, we use Correction Rate to measure whether the model follows corrective guidance and successfully produces the correct result for the targeted step. $CR=\frac{\text{Count of Effective correction Corrected}}{\text{Counts of Attempts}}\times100\%$.

\textbf{Token Cost.} Token cost comparison with baseline, including input and output. Note: The dialogue approach uses full, uncompressed history, causing input costs to rise quickly. The count is the average cost per token over two runs, and only includes questions answered correctly earlier than xR; questions not answered correctly by xR or in subsequent rounds are not included in the statistics.

\subsection{Case Studies}
Due to space limitations, we select two conceptual question involving safety considerations and logic reasoning as illustrative cases (Figure~\ref{fig:logic} and~\ref{fig:preference}). Additional case studies can be found in Appendix~\ref{casestudy}.

\begin{figure}[!h]
    \centering
    \includegraphics[width=1\linewidth]{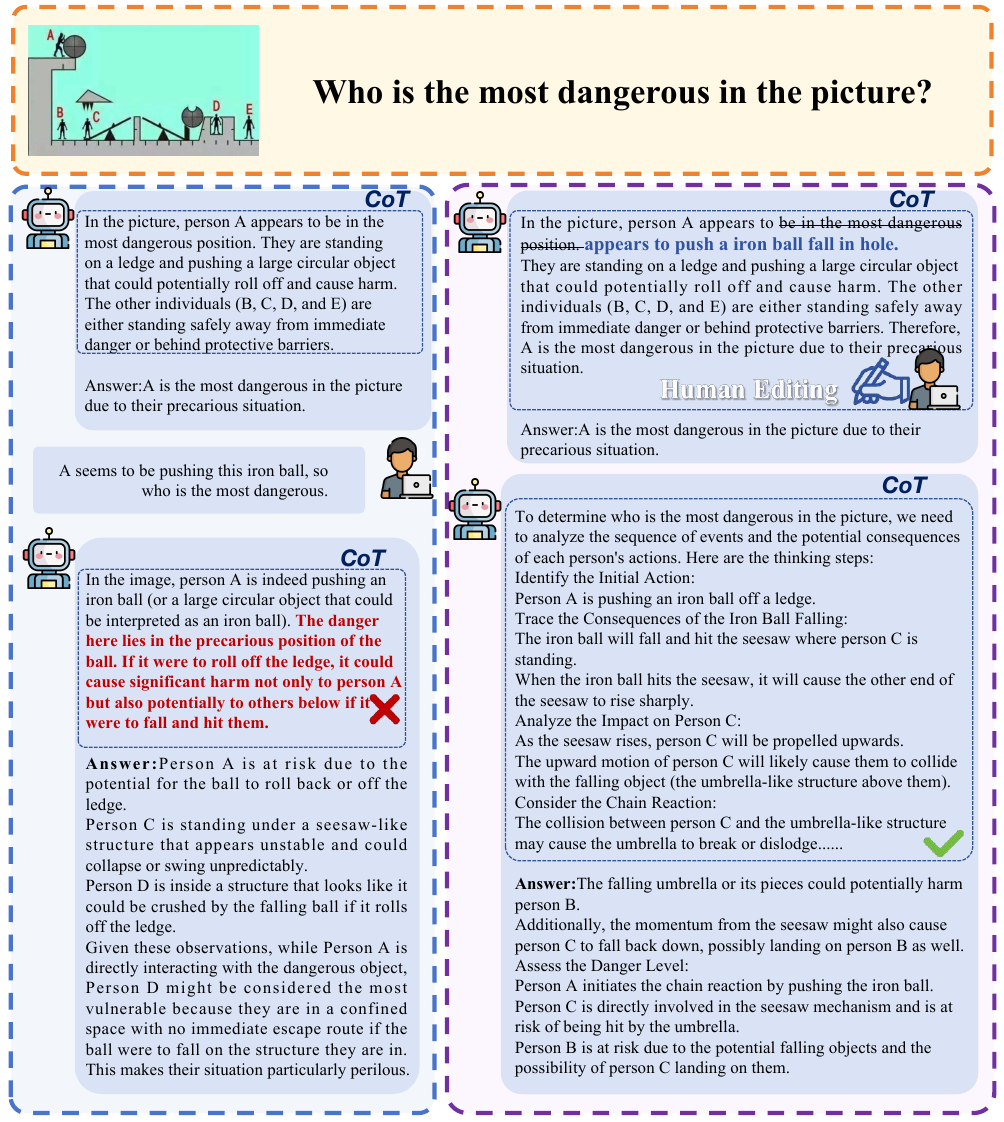}
    \caption{Case study of edit-based correction for causal reasoning. By revising only the critical reasoning span, Deep Interaction steers the model toward a response that better follows the edited rationale.}
    \label{fig:logic}
\end{figure}


\begin{figure}[!h]
    \centering
    \includegraphics[width=1\linewidth, keepaspectratio]{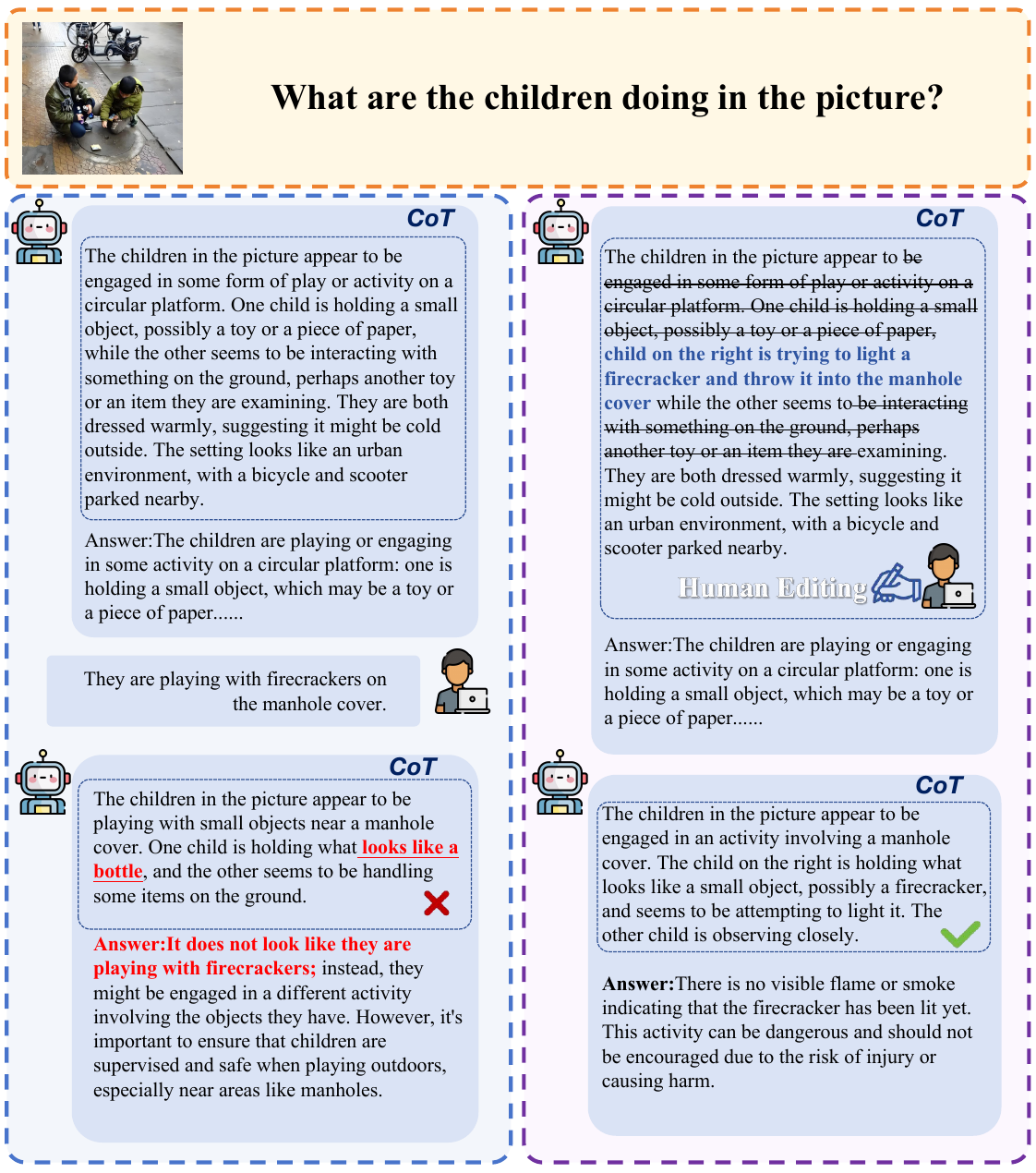}
    \caption{Case study of scenario correction involving visual misinterpretation and safety reasoning. Editing the key semantic span helps the regenerated response better align with the intended scene understanding and question focus.}
    \label{fig:preference}
    
\end{figure}

\subsection{Quantitative Results and Analysis}
Quantitative experiments primarily employ multimodal data to increase task difficulty and enable a more comprehensive evaluation of our method's performance on complex reasoning problems involving images. Baseline models, operating without the Long Thinking mechanism, achieve a direct pass rate of 93.2\%±1\% on the ScienceQA benchmark in preliminary tests. However, their performance drops over 4\% when questions involve images. Therefore, using more challenging multimodal data allows for a clearer distinction and more meaningful comparison of model performance. Correcting misinterpretations in image understanding is a key objective of our work.

\textbf{Automated Evaluation and Human Alignment.} 
Our evaluation originated with manual testing of 400 STEM problems to establish a rigorous proof of concept. Facing the immense workload of broader benchmarks like STEM20K, we developed QAgent, an automated framework that utilizes reference solutions to simulate human corrective intervention. This transition allowed us to scale our experiments efficiently while preserving the integrity of iterative reasoning evaluation.

To confirm QAgent's reliability as a high-fidelity proxy, we conducted an alignment study with students from top 100 QS universities. On 200 of the most difficult problems where the model initially failed, QAgent’s correction paths and success rates closely mirrored those of human experts as shown in Table \ref{human_alignment}. This strong correlation validates QAgent as a dependable automated evaluator, providing the necessary credibility for the large-scale performance improvements presented in Table \ref{table1-overall}.

\begin{table}[htbp]
\centering
\begin{minipage}{\columnwidth}
    \caption{Comparison of pass within different rounds.}
    \centering
    \tiny
    \begin{tabular}{l@{\hskip 0.12em}|cccc}  
        \toprule
        \textbf{Within} & \textbf{Within 1R} & \textbf{Within 2R} & \textbf{Within 3R} & \textbf{Within 4R} \\
        \midrule
        QAgent & 0 & 50 & 100 & 150 \\
        Human & 10 & 58 & 112 & 172 \\
        Gap & - & 0.16 & 0.12 & 0.147 \\
        \bottomrule
    \end{tabular}
\label{human_alignment}
\end{minipage}
\end{table}

\begin{table}
\centering
\begin{minipage}{\columnwidth}
    \caption{Comparison of pass rates across rounds.}
    \centering
    \tiny
    \begin{tabular}{l@{\hskip 0.20em}|c@{\hskip 0.30em}|c@{\hskip 0.40em}c@{\hskip 0.40em}c@{\hskip 0.40em}c}  
        \toprule
        \textbf{ScienceQA} & Pass directly & Within 1R & Within 2R & Within 3R & Within 4R \\
        \midrule
        Dialog based & 89.12\% & 94.31\% & 96.45\% & 97.27\% & 98.05\%\\
         \midrule
        Ours & - & 97.10\% & 97.93\% & 98.59\% & 99.05\%\\

        \midrule
        \midrule
        \textbf{STEM20K} & Pass directly & Within 1R & Within 2R & Within 3R & Within 4R \\
        \midrule
        Dialog based & 54.15\% & 65.18\% & 72.93\% & 78.72\% & 80.35\%\\
         \midrule
        Ours & - & 74.89\% & 79.27\% & 81.52\% & 86.69\%\\

       \rowcolor{gray!15} Ours-Expert & - & 82.36\% & 86.45\% & 89.55\% & 92.41\% \\

        \midrule
        \midrule
        \textbf{Gaokao-MM} & Pass directly & Within 1R & Within 2R & Within 3R & Within 4R \\
        \midrule
        Dialog based & 67.96\% & 80.76\% & 85.61\% & 87.30\% & 88.85\%\\
         \midrule
        Ours & - & 86.98\% & 88.91\% & 91.25\% & 93.42\%\\

       \rowcolor{gray!15} Ours-Expert & - & 90.56\% & 92.27\% & 94.74\% & 96.60\% \\
        
        \bottomrule
        \toprule
        \color{magenta}\textbf{LogicQA} & \color{black}Pass directly & Within 1R & Within 2R & Within 3R & Within 4R \\
        \midrule
        Dialog based & 81.67\% & 91.76\% & 94.81\% & 95.70\% & 96.75\%\\
        \midrule
        Ours & - & 94.36\% & 96.47\% & 97.12\% & 98.04\%\\
        \rowcolor{gray!15} Ours-Expert & - & 97.25\% & 98.46\% & 99.14\% & 99.57\% \\
        \bottomrule
    \end{tabular}
\label{table1-overall}
\end{minipage}
\end{table}

\textbf{Overall Performance.} As Table \ref{table1-overall} shows, we compare the performance of our proposed method against existing dialogue based correction approaches on the STEM challenging dataset using pass rates. ScienceQA is relatively simpler, typically requiring reasoning within two steps and involving minimal calculation, expert mode is not employed here. The results show that our method significantly outperforms existing dialogue-based correction methods.

\begin{figure}
    \centering
    \includegraphics[width=1\linewidth]{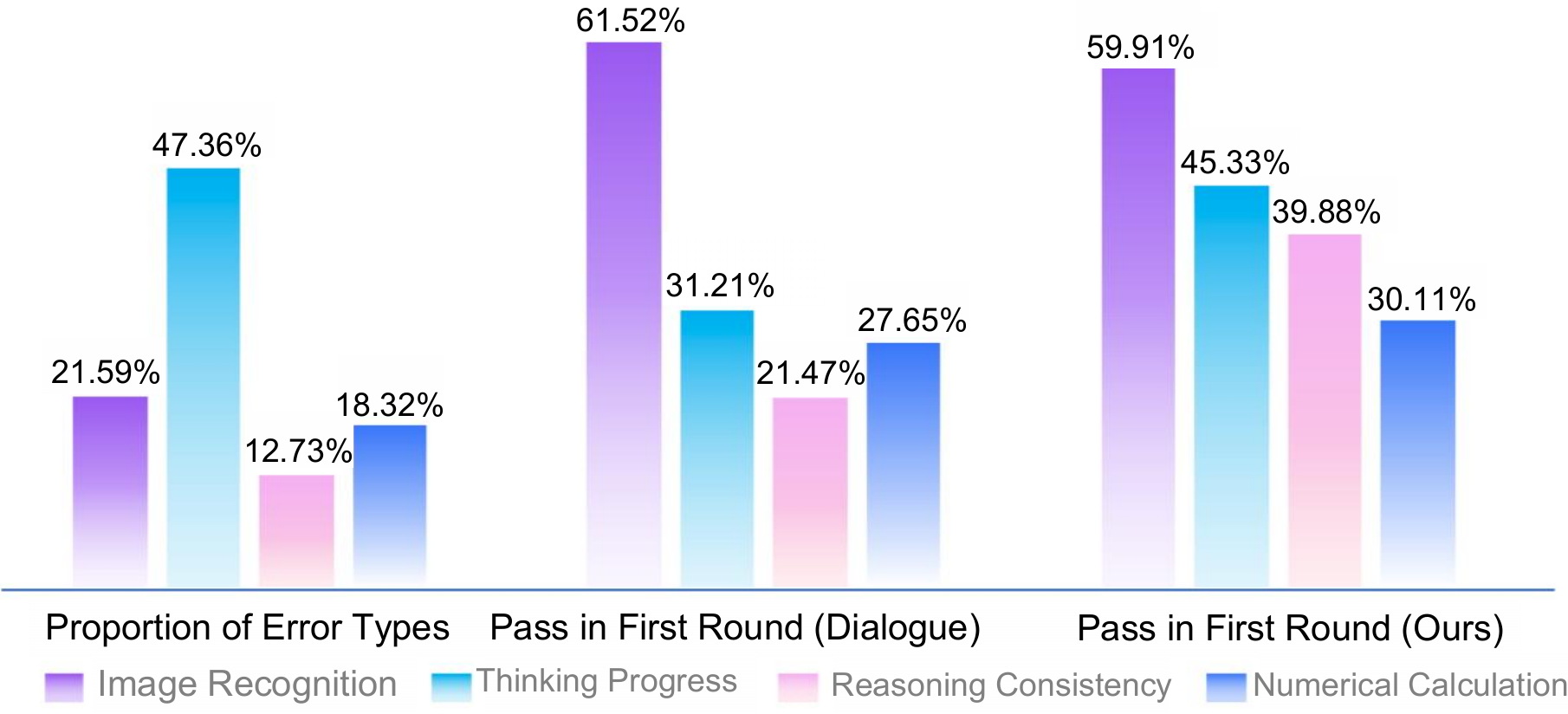}
    \caption{Comparison of one round interaction pass rate changes in different error types.}
    \label{fig:errortype}
\end{figure}

Figure \ref{fig:errortype} indicates multi-modal language models (MLLMs) still lack robust visual semantic understanding. Both dialogue based methods and our approach exhibit limited effectiveness in correcting information loss and misinterpretations arising from image recognition. In the future, enhancing the image comprehension capabilities of MLLMs will remain a key focus for improvement. In summary, whether looking at the first-round correction pass rate or the correction rate, the relative increase rate has averaged more than 25\%.

\textbf{Different LLMs.} Experimental results show that our method remains effective across various models compared to dialogue-based approaches. Models also exhibit differing levels of sensitivity to correction instructions. Although Gemini demonstrates moderate baseline capabilities, it shows strong adherence to correction instructions, as reflected in the Correction Rate (CR) in Table \ref{table2-repeat}.

\setlength{\intextsep}{3pt}
\setlength{\textfloatsep}{3pt}
\begin{table}[H]
\caption{Comparison of methods on different LLMs. GPT4o, Gemini-2.0, Claude-3.7 are employed.}
    \centering
    \scriptsize
    \begin{tabular}{c@{\hskip 0.17em}c|@{\hskip 0.17em}c@{\hskip 0.17em}|cc|@{\hskip 0.17em}c@{\hskip 0.17em}}  
        \toprule
         & Method & Pass Directly& Within 1R &Within 2R & CR  \\
        \midrule
        \multirow{2}{*}{GPT} 
        & Dialogue &\multirow{2}{*}{49.69\%}  &69.97\% & 77.70\% & 67.36\% \\
        & Ours & &73.99\% & 84.98\% & 78.93\% \\
        \midrule
        \multirow{2}{*}{Gemini} 
        & Dialogue &\multirow{2}{*}{52.01\%} &77.24\% & 85.29\% & 78.81\% \\
        & Ours & &83.77\% & 89.47\% & 90.06\% \\
        \midrule
        \multirow{2}{*}{Claude} 
        & Dialogue &\multirow{2}{*}{58.82\%} &80.50\% & 87.46\% & 70.86\% \\
        & Ours & &84.67\% & 90.56\% & 83.45\% \\
        \bottomrule
    \end{tabular}
\label{table2-repeat}
\end{table}


        
\begin{figure}
    \centering
    \includegraphics[width=0.95\linewidth]{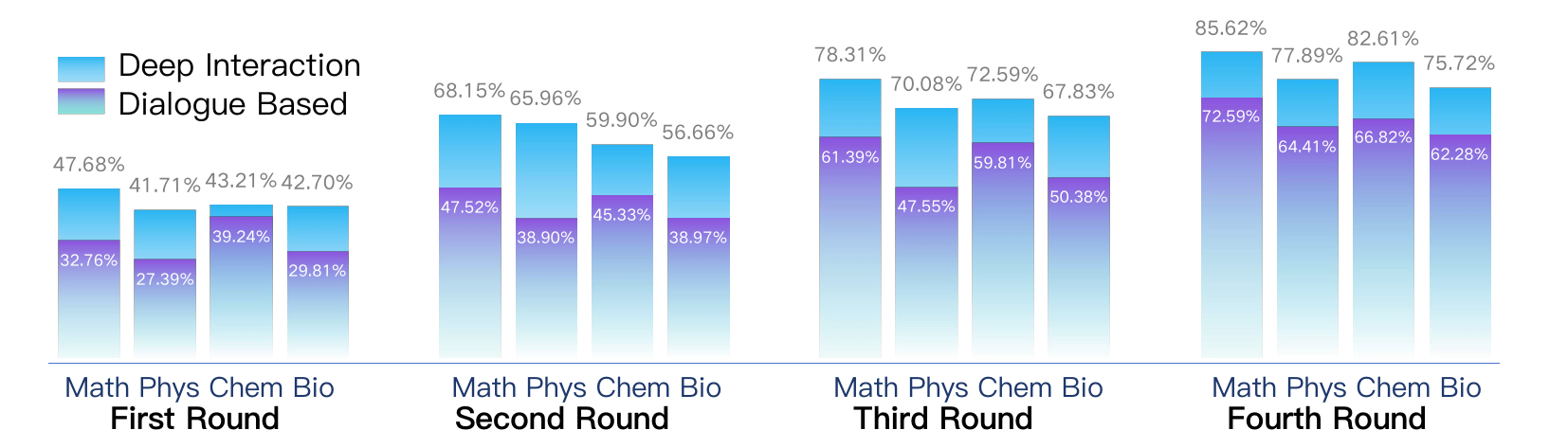}
    \caption{Comparison of correction sensitivity for different error types. Only first round of correction for each question type are taken into consideration. Because the error type of new round response of may change.}
    \label{fig:diffsub}
\end{figure}

\textbf{Different Subjects Performance.} Our analysis of error correction sensitivity reveals that Physics and Biology perform poorly across both dialogue-based and deep interaction methods. Although the correction rate for Biology appears relatively high in Figure \ref{fig:gaokaommsubcorr}, this stems from a small sample size. Detailed data can be found in Table \ref{tab:SCIENCECEE} and Figure \ref{fig:diffsub}.

\begin{minipage}[b]{0.43\columnwidth} 
    \centering
    \scriptsize
    \captionof{table}{Subject correction rate in STEM20K.}
    \begin{tabular}{l@{\hskip 0.15em}|@{\hskip 0.12em}cc}
        \toprule
        & Dialogue & Ours \\
        \midrule
        \rowcolor{gray!15} \tiny{Math} & 55.18\% & 72.20\% \\
        \rowcolor{gray!15} \tiny{Phys.} & 50.15\% & 67.37\% \\
        \rowcolor{gray!15} \tiny{Chem.} & 48.62\% & 71.17\% \\
        \rowcolor{gray!15} \tiny{Bio.} & 45.41\% & 62.57\% \\
        \bottomrule
    \end{tabular}
    
    \label{tab:SCIENCECEE}
\end{minipage}
\hfill 
\begin{minipage}[c]{0.49\columnwidth} 
    
    \centering
    
    \includegraphics[width=\linewidth, height=2.8cm]{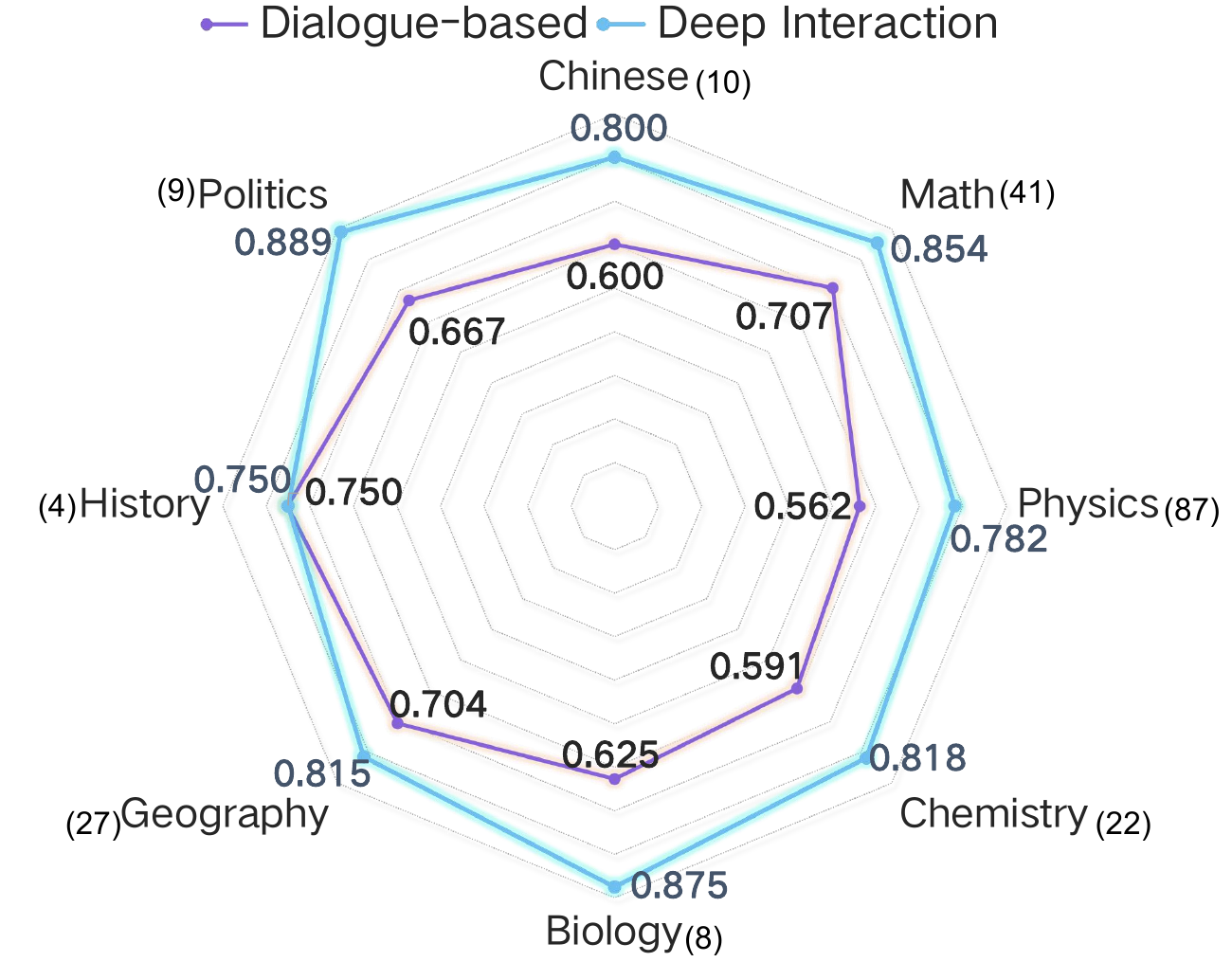} 
    \captionof{figure}{Subject correction rate in Gaokao-MM.}
    \label{fig:gaokaommsubcorr}
\end{minipage}

\textbf{Self-correction Method.} We also compare our method against Recursive Chain-of-Thought (RCoT) \cite{xue2023rcot} , a baseline that relies on reverse-thinking self-correction. Experimental results demonstrate that our approach achieves significantly higher accuracy and logical convergence across interaction rounds, yielding a correction success rate 74.8\% higher than the baseline while optimizing token efficiency. Comprehensive details are provided in Appendix~\ref{app:baseline_details}.

\begin{figure}
    \centering
    \includegraphics[width=0.95\linewidth, height=3.2cm]{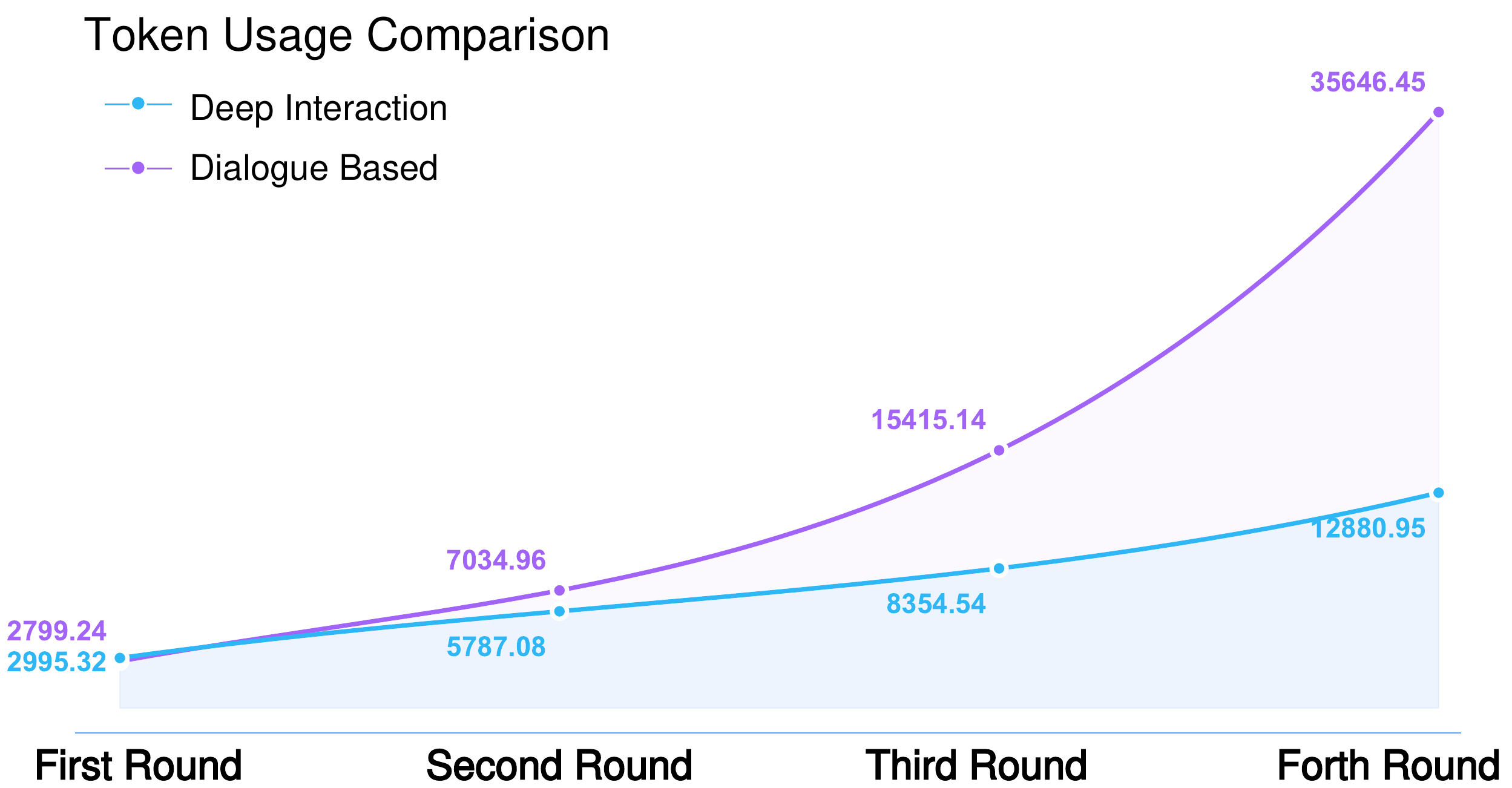}
    \caption{Comparison of token cost.}
    \label{fig:tokencost}
\end{figure}

\textbf{Token Cost.} Existing dialogue-based correction relies on conversation history, but our foundation model calls use fresh sessions without history optimization, so including full history has limited impact. Moreover, multi-turn interactions in dialogue greatly increase input tokens, details in Figure \ref{fig:tokencost}. As shown, after two rounds, the token cost increase for  corrections rises dramatically from 21.5\%. Results show a over 40\% reduction in average token cost without historical context optimization.

\subsection{Generalization Across Model Scales}
\label{sec:model-scale-generalization}

To address the concern that our main results rely on strong proprietary models, we further evaluate our method on smaller open models from the same family: \textbf{Qwen2.5-VL-7B}, \textbf{32B}, and \textbf{72B}. We report \textit{Direct Pass}, \textit{Within 1R}, \textit{Within 2R}, and \textit{CR}. We use \textbf{Gaokao-MM} for this analysis, since it provides a more informative comparison across scales. As shown in Table~\ref{tab:model-scale-generalization}, our method consistently outperforms the dialogue-based baseline at all scales. The gains are smaller on 7B, suggesting that model capability remains a bottleneck, but the overall trend shows that our method is not limited to frontier-scale models.

\begin{table}[t]
\centering

\setlength{\tabcolsep}{5pt}
\caption{Generalization across Qwen2.5-VL model scales on Gaokao-MM. Direct, 1R, and 2R denote pass rates before correction and after up to one/two correction rounds; CR denotes correction rate.}
\label{tab:model-scale-generalization}
\scriptsize
\begin{tabular}{lcccc}

\toprule
\textbf{Model / Method} & \textbf{Direct} & \textbf{1R} & \textbf{2R} & \textbf{CR} \\
\midrule
Qwen2.5-VL-7B (Dialogue)  & 19.8 & 40.6 & 56.5 & 68.4 \\
Qwen2.5-VL-7B (\textsc{Ours}) & 19.8 & \textbf{41.8} & \textbf{59.2} & \textbf{70.1} \\
\addlinespace[2pt]
Qwen2.5-VL-32B (Dialogue)  & 34.7 & 48.3 & 63.5 & 72.0 \\
Qwen2.5-VL-32B (\textsc{Ours}) & 34.7 & \textbf{50.8} & \textbf{68.3} & \textbf{79.6} \\
\addlinespace[2pt]
Qwen2.5-VL-72B (Dialogue)  & 42.4 & 65.5 & 68.6 & 72.5 \\
Qwen2.5-VL-72B (\textsc{Ours}) & 42.4 & \textbf{69.7} & \textbf{74.3} & \textbf{78.2} \\
\bottomrule
\end{tabular}
\end{table}

\subsection{Compare with Other Methods}
We compare Deep Interaction with RCoT \cite{xue2023rcot}, Self-Refine, and Interactive Reasoning; see Appendix~\ref{app:baseline_details} for details. Table~\ref{table:di_vs_baselines_performance} shows that Deep Interaction achieves the best Round-5 accuracy of 96.8\%, exceeding the strongest baseline by 0.7\%.

\begin{table}[H]
\caption{Performance Comparison among RCoT, Self-Refine, Interactive Reasoning, and Deep Interaction.}
\centering
\scriptsize
\setlength{\tabcolsep}{3pt}
\begin{tabular}{l ccccc}
\toprule
\textbf{Method} & \textbf{Round 1} & \textbf{Round 2} & \textbf{Round 3} & \textbf{Round 4} & \textbf{Round 5} \\
\midrule
RCoT & \multirow{4}{*}{81.60\%} & 85.2\% & 85.8\% & 86.3\% & 87.3\% \\
Self-Refine &  & 88.7\% & 89.2\% & 90.1\% & 90.6\% \\
Interactive Reasoning &  & 90.5\% & 94.2\% & \textbf{95.7\%} & 96.1\% \\
\textbf{Deep Interaction} &  & \textbf{91.8\%} & \textbf{94.8\%} & 95.6\% & \textbf{96.8\%} \\
\bottomrule
\end{tabular}
\label{table:di_vs_baselines_performance}
\end{table}

\subsection{Ablation Studies}
We conduct ablation studies on Gaokao-MM to evaluate the editable response mechanism, Text Processing, and the CoT Reprompter. Detailed settings and metric definitions are provided in Appendix~\ref{app:ablation_details}. As shown in Table~\ref{tableabla}, replacing editable responses with 3-round history or removing either module leads to clear performance degradation, confirming that direct CoT editing, structured text processing, and targeted re-prompting are all necessary for effective correction.
\begin{table}[h]
\caption{Ablation results for each module.}
    \centering
    \tiny
    \begin{tabular}{l@{\hskip 0.12em}l@{\hskip 0.15em}|@{\hskip 0.3em}c@{\hskip 0.35em}c|cc}  
        \toprule
         & & With 1R$_{(\uparrow)}$ & Correction$_{(\uparrow)}$ & Token$_{(\downarrow)}$ & Latency$_{(\downarrow)}$  \\
        \midrule
        
        w/ &3R History & -8.97\% & -10.89\% & +1.27\%$^*$ & +11.91\% \\

        w/o &Text Processing & -15.27\% & -23.31\% & -7.87\% & -3.12\%  \\

        w/o &CoT Reprompter & -12.49\% & -17.57\% & +5.45\% & -12.87\%  \\
        
        \bottomrule
    \end{tabular}
\label{tableabla}
\end{table}

\section{Conclusion}
In this work, we propose \textit{Deep Interaction}, a direct-editing paradigm for efficiently revising LLM reasoning. Experiments show that it reduces redundant dialogue, improves correction success by over 25\% on STEM and logical reasoning benchmarks, and lowers token overhead by about 40\%. It also enhances instruction following in open-ended tasks, suggesting its potential for human-AI co-creation.

\section*{Limitations}

\subsection*{Dependency on User Correction Quality}
Our framework assumes that users can identify at least part of the erroneous reasoning and provide meaningful edits. Therefore, its effectiveness depends on the user's ability to detect flawed reasoning and supply useful corrections. This assumption is more realistic for users with sufficient task understanding, such as expert or semi-expert users, than for general users.

\subsection*{Sensitivity to Erroneous Human Guidance}
While Deep Interaction significantly improves the model's ability to follow human instruction in non-fixed answer scenarios, this high degree of controllability creates a risk of error propagation. If a user provides an incorrect correction, the model may faithfully follow the flawed guidance and rationalize the resulting hallucination.

\subsection*{Boundary of Collaborative Correction}
This paradigm serves as a collaborative mechanism rather than a replacement for a model's intrinsic reasoning or fully automated error-detection systems. For users unable to provide cognitively sound feedback, the burden of correction falls outside the scope of this interaction framework and must instead be addressed by the model's baseline capabilities or specialized automated agents.

\subsection*{Dependence on Exposed Reasoning Traces}
Our framework currently relies on models or systems that expose editable intermediate reasoning traces, such as LLMs with visible Chain-of-Thought outputs or web/agent systems that provide intermediate decision trajectories. In settings where only final answers are available, the method is less directly applicable. In addition, part of our large-scale evaluation relies on LLM-based automatic judging, which may introduce measurement bias despite our calibration efforts.

\bibliography{custom}
\newpage
\appendix

\section*{Appendix}
\label{sec:appendix}

\section{Supplementary Related Work}
Integrating human feedback (HF) with Chain-of-Thought (CoT) reasoning aims to improve the correctness, reliability, and human alignment of LLM-generated reasoning paths. While CoT externalizes reasoning~\cite{yu2023towards6}, it often contains logical or factual errors~\cite{12xia2024beyond}, motivating the use of HF for correction.

Feedback can target either final outcomes or intermediate steps. Outcome supervision, based on answer correctness, is easier to collect but may overlook flawed reasoning~\cite{1li2025enhancing}. In contrast, process supervision evaluates individual steps, enabling finer-grained correction and yielding better performance on complex tasks~\cite{81lightman2023let}, albeit at higher annotation cost~\cite{27sun2025efficient}. Extensions include generative PRMs like ThinkPRM~\cite{85khalifa2025process}, hierarchical models (HRMs) assessing multi-step coherence~\cite{29wang2025towards}, and generative RMs (GenRMs) that produce CoT rationales for more interpretable rewards~\cite{87guo2025reward, 88hong2025think}.

To reduce annotation burden, methods like RLTHF use targeted human feedback on critical samples~\cite{19ouyang2022training}, while approaches such as DCoT encourage intra-inference refinement via diverse CoT generation~\cite{91puerto2025fine}. Despite progress, challenges remain in reward model robustness and alignment quality~\cite{17kaufmann2024survey}. Recent work extends feedback beyond correctness to meta-properties of CoT, such as Reasoning Verbosity (RV) and Cognitive Difficulty (CD), as formalized in OmniThought~\cite{83cai2025reasoning}. These dimensions promote reasoning that is not only correct but also contextually appropriate and cognitively aligned, moving toward optimal reasoning. Concurrently, the cost of manual annotation~\cite{4yao2025unveiling} has driven human-AI collaborative frameworks: AI-generated CoTs are refined by humans~\cite{96cohn2025cotal}, critical samples are prioritized for review (e.g., RLTHF as mentioned above), or AI acts as a teacher/judge with human oversight (e.g., RGF~\cite{97diallo2025rule}, LLM-as-a-Judge~\cite{101son2024llm}). Such symbiotic approaches balance scalability and precision in CoT refinement.

Notably, some prior work has explored human involvement to improve CoT quality. \cite{AAcai2023human} proposes generating multiple outputs and presenting high confidence answers to users, a strategy conceptually similar to beam search \cite{xie2023self}. Final outputs are then refined via human feedback, but this approach essentially relies on post-hoc dialogue based correction, which becomes prohibitively costly for multi-step problems requiring step-by-step human review and complex implementation. B introduces a step-wise pause then confirm framework, a strategy we independently considered during design. However, as our investigation reveals, such methods are cumbersome in practice due to the overhead of iterative pauses and feedback. Moreover, \cite{BByoo2025co} lacks detailed implementation descriptions and empirical validation. In contrast, our method modifies the model’s initial response in a single round of interaction. We ground our design in user studies with both experts and non-experts, ensuring practical efficiency. Furthermore, we contribute theoretical innovations and provide preliminary justification for the feasibility of our approach.

\section{Acquiring Fine-Tuning Data for the CoT Reprompter}
\label{finetunedata}
By accumulating human-corrected data, we obtain a large number of [CoT edited by human] with [fully correct CoTs and final answers]. Through the following process, we further construct fine-tuning data as multiple pairs of [CoT edited by human]–[semantically equivalent minimal CoT].
\setcounter{AlgoLine}{0}
\begin{algorithm}[t]
\caption{Fine-tune Dataset Process}\label{alg:process}
\small
\KwIn{Initial edited CoT $\mathcal{C}_r'$,\; Right answer $(C, A)\color{green}{\usym{2714}}$}
\KwOut{The shortest valid simplified hint from $FC_{r}$}

\tcp{Initialize $\mathcal{C}_r'$ as a Query sample}
$Q^s \leftarrow \mathcal{C}_r'$\; 

$fail\_count \leftarrow 0$\; \tcp*[r]{Initialize fail counter}
\tcp{Set maximum allowed failures}
$N \leftarrow 3$\; 

\While{$\text{fail\_count} < N$}{
    \BlankLine
    \tcp{Simplify question by LLM}
    $Q^{s}{'} \leftarrow \text{LLM}(Q^{s},\text{simplify prompt})$\; 

    \tcp{Answer questions by LLM}
    $(\mathcal{C}, \mathcal{A}) \leftarrow \text{LLM}(\mathcal{Q}_r,Q^{s}{'})$\; 
    
    \If{$(\mathcal{C}, \mathcal{A}), (C, A)) = \text{True}$}{
        \tcp{Update it with new version}
        $Q^s \leftarrow Q^{s}{'}$\;

        $fail\_count \leftarrow 0$\; \tcp{Reset fail counter}
    }\Else{
        \tcp{Increase fail counter}
        $fail\_count \leftarrow fail\_count + 1$\; 
    }
}
\Return $Q^s$\; \tcp*[r]{Get final simplified CoT}
\end{algorithm}

\section{Experimental Description}
\subsection{Statistical Explanation of Evaluation Metrics}
Due to the enormous volume of our questions, we could not perform five rounds of validation; therefore, we conducted two rounds, defining that a question is counted as passed if it passes at least once. Meanwhile, we carried out full human verification on ScienceQA, while using machines for comprehensive evaluation on other datasets, and invited students from top 100 QS universities to cross-check the answers against the standard answers, ensuring the effectiveness of machine evaluation. We randomly sampled 100 questions from each subject in STEM20K for manual review. The results were largely consistent with those of QAgent: the recall rates are shown in the Table \ref{tab1} below. We can thus conclude that QAgent's capabilities are close to real-world performance.

\begin{table}[htbp]
\centering
\begin{minipage}{\columnwidth}
    \caption{Comparison of error rates across subjects.}
    \centering
    \tiny
    \begin{tabular}{l@{\hskip 0.12em}|cccc}  
        \toprule
        \textbf{Subject} & \textbf{Mathematics} & \textbf{Physics} & \textbf{Chemistry} & \textbf{Biology} \\
        \midrule
        Manually Confirmed Errors & 52 & 44 & 48 & 45 \\
        QAgent Output Errors & 53 & 46 & 45 & 45 \\
        Actual Hit Errors & 49 & 42 & 43 & 43 \\
        \midrule
        Recall Rate & 0.942 & 0.954 & 0.895 & 0.956 \\
        False Positive Rate & 0.077 & 0.091 & 0.042 & 0.044 \\
        \bottomrule
    \end{tabular}
    \label{tab1}
\end{minipage}
\end{table}

\subsection{Test pipeline and QAgent}
\label{QAGENT}
\subsubsection{Methodology}
To address the research questions, we developed a multimodal Question-Answering Agent (QAgent). The core function of this agent is to facilitate a multi-turn conversational process with a target Large Language Model (LLM), using another LLM as a judge to evaluate the responses. The process begins with a multimodal question, consisting of both text and an image, which is presented to the target LLM. The agent then takes the answer provided by the target LLM and compares it against a standard answer, with a particular focus on the consistency of the chain-of-thought process. This comparison is performed by a separate "judge" LLM. If the target LLM's answer is deemed incorrect, the judge LLM generates feedback, which is then used to construct a new prompt for the target LLM in the subsequent turn. This iterative process continues until the target LLM provides a correct answer or a predefined maximum number of rounds is reached.
\subsubsection{QAgent}

The QAgent is a system designed to automate the process of evaluating and improving the responses of a target LLM for multimodal questions. It orchestrates the interaction between the target LLM and a judge LLM. The agent is responsible for:
1. Receiving multimodal questions (text and image) and the corresponding standard answers.
2. Sending the question to the target LLM for a response.
3. Utilizing a judge LLM to evaluate the target LLM's answer against the standard answer, with an emphasis on the reasoning process.
4. Generating corrective feedback through the judge LLM if the answer is incorrect.
5.Managing the multi-turn iterative process, continuing until a correct answer is obtained or a maximum number of attempts is made.

\subsubsection*{Implementation Pseudocode} 

\setcounter{AlgoLine}{0}
\begin{algorithm}[t]
\caption{Multimodal QAgent with Iterative Feedback}
\label{alg:qa_agent}
\small
\KwIn{A multimodal question $Q$, an image $I$, a standard answer $A_s$, and max rounds $R_{\max}$}
\KwOut{A detailed experiment history $H_{\exp}$}

$H_{\exp} \leftarrow \emptyset$,\quad $round \leftarrow 0$,\quad $is\_correct \leftarrow \text{false}$,\quad $feedback \leftarrow \text{""}$\;

\While{$round < R_{\max}$ \textbf{and} $is\_correct = \text{false}$}{
    $round \leftarrow round + 1$\;
    \eIf{$round = 1$}{
        $P_{\text{prompt}} \leftarrow Q$\;
    }{
        $P_{\text{prompt}} \leftarrow Q + \text{" Feedback: "} + feedback$\;
    }
    
    $A_{\text{target}} \leftarrow \text{CallTargetLLM}(P_{\text{prompt}}, I)$\tcp*{Get answer from target model}
    
    $J_{\text{result}} \leftarrow \text{CallJudgeLLM}(Q, A_s, A_{\text{target}})$\tcp*{Evaluate with judge model}
    
    $(is\_correct,\, feedback) \leftarrow \text{ParseJSON}(J_{\text{result}})$\tcp*{Extract correctness and feedback}
    
    append $(round,\, P_{\text{prompt}},\, A_{\text{target}},\, is\_correct,\, feedback)$ to $H_{\exp}$\;
}
\Return $H_{\exp}$\;
\end{algorithm}
\subsection{Experimental Insights and Future Work}
Table \ref{tab:stability-generalization} further shows over 90\% accuracy on repeated questions using the final correct CoT, compared to about 60\% when using full dialogue input. Our approach also generalizes well to modified parameters, question formats, and image changes, demonstrating both consistency and strong generalization.

Considering error correction and preference alignment, we plan to enhance our system with features for individual users, including error memory and preference alignment. Our goal is to ensure that questions previously answered incorrectly and subsequently corrected will not be repeated. For similar problems, the system will enable users to learn from one example and apply the knowledge to others. Additionally, the system will remember user preferences, such as whether they prioritize detailed derivation processes or conceptual explanations, and for coding tasks, whether they prefer concise code or clear comments.

\begin{table*}[h]
\caption{Stability and generalization ability}
    \centering
    \scriptsize
    \begin{tabular}{l|c|ccc}  
        \toprule
         & Repeat. & Varying Query Parameters & Query Harder & Modifying Query Figure Content  \\
        \midrule
        Dialogue based & 61.3\% & 27.2\% & 18.9\% & 30.7\% \\
         \midrule
        Deep Interaction & 92.0\% & 58.8\% & 59.2\% & 61.5\% \\

        E. Deep Interaction & 95.5\% & 43.5\% & 60.4\% & 60.5\%  \\
        
        \bottomrule
    \end{tabular}
\label{tab:stability-generalization}
\end{table*}

\subsection{Quantitative Comparison with RCoT}
\label{app:baseline_details}
To quantitatively evaluate Deep Interaction on complex reasoning error correction, we compare it with representative correction baselines, including RCoT \cite{xue2023rcot}, Self-Refine, and Interactive Reasoning. RCoT relies on reverse reasoning to detect inconsistencies from the answer back to the question, while Self-Refine and Interactive Reasoning use iterative self-feedback or step-wise interaction to improve responses. Although these methods can correct local or easily detectable errors, they may still struggle when the model follows a seemingly coherent but flawed reasoning trajectory.

As shown in Table \ref{table:di_vs_baselines_performance}, all methods start from the same first-round performance of 81.6\%, but their improvement trends diverge in later rounds. RCoT quickly plateaus after several rounds, reaching 87.3\% in Round 5, while Self-Refine improves to 90.6\%. Interactive Reasoning achieves stronger performance with 96.1\%, showing the value of external guidance. Deep Interaction achieves the best final result of 96.8\%, outperforming RCoT by 9.5\% and Self-Refine by 6.2\% in absolute accuracy. This suggests that directly editing erroneous CoT spans can more effectively break the model's incorrect reasoning trajectory than self-correction or conventional interaction.

\subsection{Ablation Study Details}
\label{app:ablation_details}

We conduct ablation studies on the Gaokao-MM dataset to evaluate the contribution of each component in Deep Interaction. The editable response mechanism is compared with a variant that provides a 3-round relevant dialogue history, where additional historical context is exposed to the model instead of directly editing the erroneous reasoning span. We also remove Text Processing and the CoT Reprompter to assess their individual effects.

In Table~\ref{tableabla}, With 1R and Correction denote the performance changes in first-round interaction and correction accuracy, respectively. Token and Latency report the relative changes in token consumption and response time. The symbols $\uparrow$ and $\downarrow$ indicate whether a higher or lower value is preferred for each metric. For performance metrics, negative values indicate degradation compared with the full method. For cost-related metrics, positive values indicate increased cost, while negative values indicate reduced cost. The superscript $^*$ indicates that only output-token changes are counted, excluding the increase in input tokens.

The results show that simply providing more historical dialogue is not sufficient for effective correction. Although 3-round history provides more context, it reduces With 1R and Correction by 8.97\% and 10.89\%, respectively, while also increasing latency by 11.91\%. This suggests that direct editing of erroneous CoT spans is more effective than relying on longer dialogue history. Removing Text Processing causes the largest performance drop, indicating that structured and concise input is important for stable correction. Removing the CoT Reprompter also significantly weakens performance, confirming that targeted re-prompting helps the model regenerate corrected reasoning more effectively.

\section{Theoretical Analysis}
\label{analysis}
Here, we provide a brief theoretical justification for the methodology presented in the main text, aiming to demonstrate its theoretical feasibility. First, the proposed approach involves editing the Chain-of-Thought (CoT) and then re-integrating the original query with the edited CoT to form a new input. Although this modifies the input, the core content remains largely unchanged. Therefore, it is essential to perturb the new input in a way that alters its data distribution, while preserving semantic fidelity to avoid answer deviation. We find that minimal simplification. To be specific, removing non-essential, generic, or stylistic words that contribute little to the query's specificity ,and effectively enhances the model's focus on the core query and the revised CoT content.

\textbf{Theorem 1.} \textit{Any prefix, once concatenated, locks the distribution and halts posterior information gain.}

When the model's generated prefix subsequence is explicitly appended back to the original input, the posterior generation distribution remains almost unchanged. Define the input space as \( \mathcal{T} \), the set of all finite or countable token sequences. After being trained on a dataset with a predefined data distribution, a fixed-parameter LLM follows the input-output distribution as specified below. Previous studies have also provided proof that the model's next-token distribution is fixed for the same (prompt + history) prefix \cite{holtzman2019curious}.
\begin{equation}
    p_\theta(y_{1:T}\mid x)=\prod_{t=1}^{T}p_\theta(y_t\mid x,y_{<t}),\qquad x,y_t\in\mathcal V,
\end{equation}
Where the input \( x \in \mathcal{T} \) denotes the original query, and $\mathcal V$ is the vocabulary list that LLM used. Given the input \( x \), the model generates a sequence by sampling: 
\begin{equation}
     y:=y_{1:T}\sim p_\theta(\cdot\mid x).
\end{equation}
By prefix truncation, take any prefix of \( y \), and construct a new prompt as Eq. \ref{eq:a}.
\begin{equation}
    y_{\leq k}:=y_{1:k}\quad(1\le k\le T),
\end{equation}
\begin{equation}
    x':=x\circ y_{\leq k},
    \label{eq:a}
\end{equation}
Then, the original and continuing generation distribution $\mathcal{R}$ and $\mathcal{R}'$are:
\begin{equation}
    \mathcal{R}:=p_\theta(\cdot\mid x,y_{\leq k}),\qquad \mathcal{R}':=p_\theta(\cdot\mid x').
\end{equation}
Therefore, the problem reduces to proving that \( \mathcal{R} = \mathcal{R}' \) holds approximately under low-temperature setting in LLMs.
Based on the definition of Prefix Invariance \cite{radford2019language}, under the causal structure of an autoregressive $p_\theta$ model, for any length $k$ and any prefix, the following holds:
\begin{equation}
    p_\theta\bigl(y_{k+1:T}\bigm|x,y_{\leq k}\bigr)=p_\theta\bigl(y_{k+1:T}\bigm|x\circ y_{\leq k}\bigr).
\end{equation}

\textbf{\textit{Proof.}} According to the autoregressive decomposition as Eq. \ref{eqb}, for autoregressive Transformers or other architectures with causal masking, the conditional probability depends only on the historical tokens, as shown in Eq. \ref{eqc}.
\begin{equation}
    p_\theta(y_{k+1:T}\mid x,y_{\leq k})=\prod_{t=k+1}^{T}p_\theta(y_t\mid x,y_{<t}),
    \label{eqb}
\end{equation}
\begin{equation}
    p_\theta(y_t\mid \text{prefix})=p_\theta\bigl(y_t\mid \text{prefix}_{<t}\bigr).
    \label{eqc}
\end{equation}
Hence, the result follows.
\begin{equation}
     p_\theta(y_t\mid x\circ y_{\leq k},y_{k+1:t-1})=p_\theta(y_t\mid x,y_{<t}),\qquad t>k.
\end{equation}
Substituting into the product Eq. \ref{eqb}, it suffices to show that \( \mathcal{R} = \mathcal{R}' \).
However, this equality should not be interpreted as strict character-level equivalence; it can be rigorously characterized as the entropy difference or KL divergence tending to zero:
\begin{equation}
    \Delta H:=H(q')-H(q)=0, 
    \mathrm{KL}(q'\parallel q)=0.
\end{equation}

\textbf{Theorem 2.} Assume the large language model is a fixed-parameter conditional distribution $p_\theta(y\mid x)$, where $\mathcal{X,Y} \in \mathcal{T}$ lies on discrete token sequence space, $y$ denotes the token sequence space of outputs. The semantic space \( S \) is a latent variable, and there exists a semantic encoding. Based on the theory of semantic entropy correlation research \cite{farquhar2024detecting, nikitin2024kernel}, we can make the following simple inferences.
\begin{equation}
    \mathsf{sem} : x \in\mathcal{X}\to\mathcal S,\qquad \text{and}\quad \mathsf{sem}(x)=\mathsf{sem}(x')=:s.
\end{equation}
For a given semantic meaning \( s \in \mathcal{S} \), define the perturbation space.
\begin{equation}
    \mathcal X_s:=\bigl\{x\in\mathcal X\mid \mathsf{sem}(x)=s\bigr\}.
\end{equation}

Suppose the model \( p_\theta \) satisfies the following conditions:

(i) For any \( x \in \mathcal{X} \), the support of \( p_\theta(\cdot \mid x) \) does not collapse to a single point;

(ii) There exist \( x_1, x_2 \in \mathcal{X}_s \) such that  
\begin{equation}
    \mathrm{KL}\!\bigl(p_\theta(\cdot\mid x_1) \parallel p_\theta(\cdot\mid x_2)\bigr) > 0.
\end{equation}
\begin{equation}
    \Pr_{x \sim \mathcal{U}(\mathcal{X}_s)}\!\Bigl[\, \mathrm{KL}\!\bigl(p_\theta(\cdot\mid x) \parallel p_\theta(\cdot\mid x_0)\bigr) > 0 \Bigr] > 0,
\end{equation}

where \( \mathcal{U}(\mathcal{X}_s) \) denotes the uniform measure over \( \mathcal{X}_s \).

\textbf{\textit{Proof.}} First, They have a lower bound of entropy. From (i), we know that $\forall x,\; H(p_\theta(\cdot\mid x)) > 0$.

According to positivity of KL divergence, then combining with (ii):  
\begin{equation}
    \exists x_1, x_2,\; p_\theta(\cdot\mid x_1) \neq p_\theta(\cdot\mid x_2).
\end{equation}

Through semantic relation mapping:
\begin{equation}
    f: x \mapsto p_\theta(\cdot\mid x) \in \mathcal{P}(\mathcal{Y})
\end{equation}

is a continuous embedding. Since $\mathcal{X}_s$ is a finite set under the discrete topology, $f$ is trivially continuous. Therefore, the set $
   \bigl\{x \in \mathcal{X}_s \mid \mathrm{KL}(p_\theta(\cdot\mid x) \parallel p_\theta(\cdot\mid x_0)) > 0\bigr\}
$ is non-empty, and has positive measure under the uniform distribution.

Next, we assume a quantitative characterization of the perturbation effect: for any \( x \in \mathcal{X}_s \), define the output entropy difference as  
\begin{equation}
\Delta H(x; x_0) := H(p_\theta(\cdot\mid x)) - H(p_\theta(\cdot\mid x_0)).
\end{equation}

Therefore, we can establish an entropy-divergence identity:
\begin{equation}
    \begin{aligned}
    \Delta H(x;x_0)
    &= \mathbb{E}_{y\sim p_\theta(\cdot\mid x)}\!\Bigl[\ln\frac{p_\theta(y\mid x)}{p_\theta(y\mid x_0)}\Bigr] \\
    &\quad - \mathrm{KL}\!\bigl(p_\theta(\cdot\mid x)\parallel p_\theta(\cdot\mid x_0)\bigr).
    \end{aligned}
\end{equation}
where the first term on the right-hand side is the cross-entropy difference, and the second term is the KL divergence; this identity follows directly from the definition.

Let $\mu$ be an arbitrary non-degenerate prior on $\mathcal{X}_s$ (e.g., a language model prior or a data augmentation strategy). Then the expected entropy increase satisfies
\begin{equation}
\mathbb{E}_{x \sim \mu}\!\bigl[\Delta H(x; x_0)\bigr] \geq 0,
\end{equation}
and
\begin{equation}
\mathbb{E}_{x \sim \mu}\!\bigl[\mathrm{KL}(p_\theta(\cdot\mid x) \parallel p_\theta(\cdot\mid x_0))\bigr] > 0
\end{equation}
if and only if the mapping $x \mapsto p_\theta(\cdot\mid x)$ is non-constant on $\mathcal{X}_s$. Therefore, it can be said that using semantically equivalent but textually diverse queries constitutes a theoretically justifiable mechanism for activating diversity, capable of breaking the model's excessive convergence toward fixed memorized patterns.

\textbf{Theorem 3.} \textit{Advantages of Semantic Redundancy Removal.} 

Removing semantic segments that are irrelevant or non-specific to the query can enhance the performance of LLMs. Suppose the input \(x\) contains a redundant subsequence \(r\) such that \(\mathsf{sem}(x) = \mathsf{sem}(x \setminus r)\). Let \(p_\theta\) be an autoregressive model, and define  
\begin{equation}
    \Delta := \mathrm{MI}(x; y) - \mathrm{MI}(x \setminus r; y),
\end{equation}

where \(\mathrm{MI}\) denotes the mutual information between input and output.  
Since \(r\) contributes only noise entropy, \(\Delta \leq 0\); by the data processing inequality, after removing redundancy we have  
\begin{equation}
    \mathrm{MI}(x \setminus r; y) \leq \mathrm{MI}(x; y). 
\end{equation}

Therefore, the denoised input retains a higher effective information density under the same semantic constraint, thereby improving output quality.

\section{Data Accessibility}
We have included a Gradio-based lightweight viewer in the appendix, along with the original texts of the first 25 questions from each subject in STEM20K for reference. Additionally, we plan to publicly release a subset of our dataset comprising approximately one thousand multimodal questions with reference answers, pending acceptance of this work.

\section{Productization and Trial Use}
\begin{figure*}[!t]
    \centering
    \includegraphics[width=0.85\textwidth]{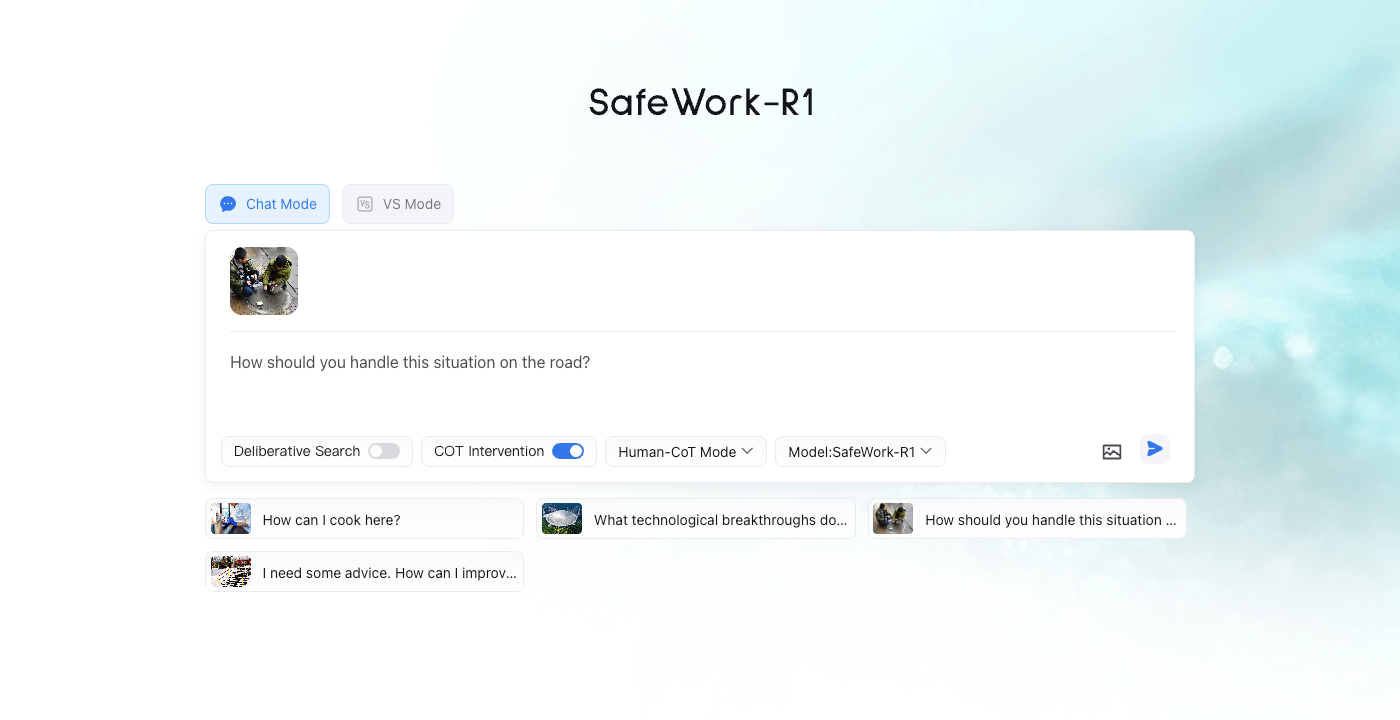}
    \caption{Human-CoT application mode deployed on SafeWork-R1~\cite{safeworkr1}.}
    \label{fig:frontpage}
\end{figure*}

As Figure \ref{fig:frontpage} shows, the Human-CoT application mode has been deployed on SafeWork-R1~\cite{safeworkr1}, a public online large model platform. The platform has passed internal testing and is currently in an invited trial phase for targeted users worldwide. The system has been adapted to integrate various large models, including our internally released models, the Qwen2.5 series, Deepseek series, Claude, and others. Screenshots are provided below. Key identifying information has been obscured to meet article review requirements.

\begin{figure*}
    \centering
    \includegraphics[width=1\linewidth]{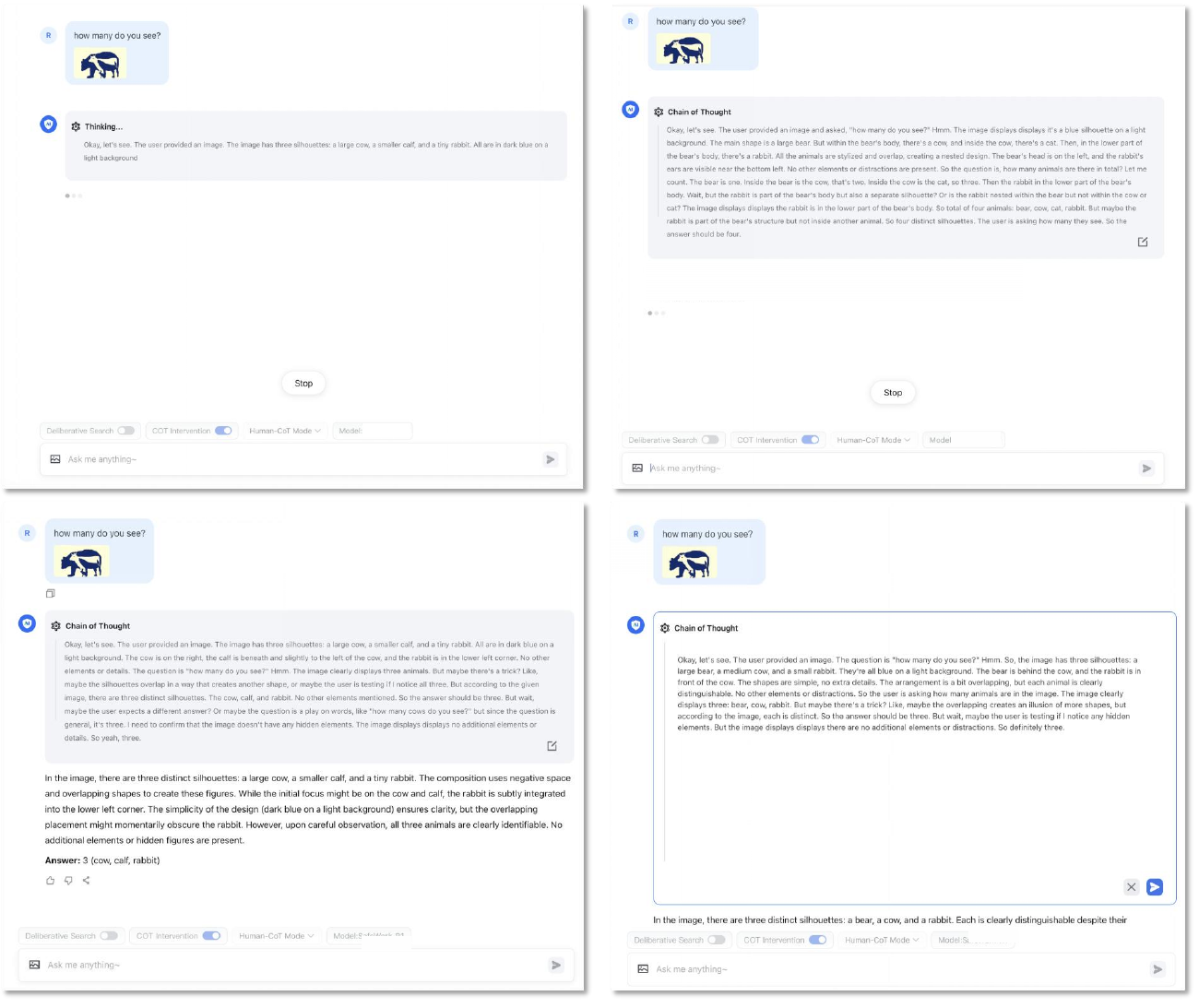}
    \caption{The first three images show the process of generating an answer after a question is entered, while the last image illustrates the response mode after clicking "CoT Edit".}
    \label{fig:4step}
\end{figure*}

\begin{figure*}
    \centering
    \includegraphics[width=1\linewidth]{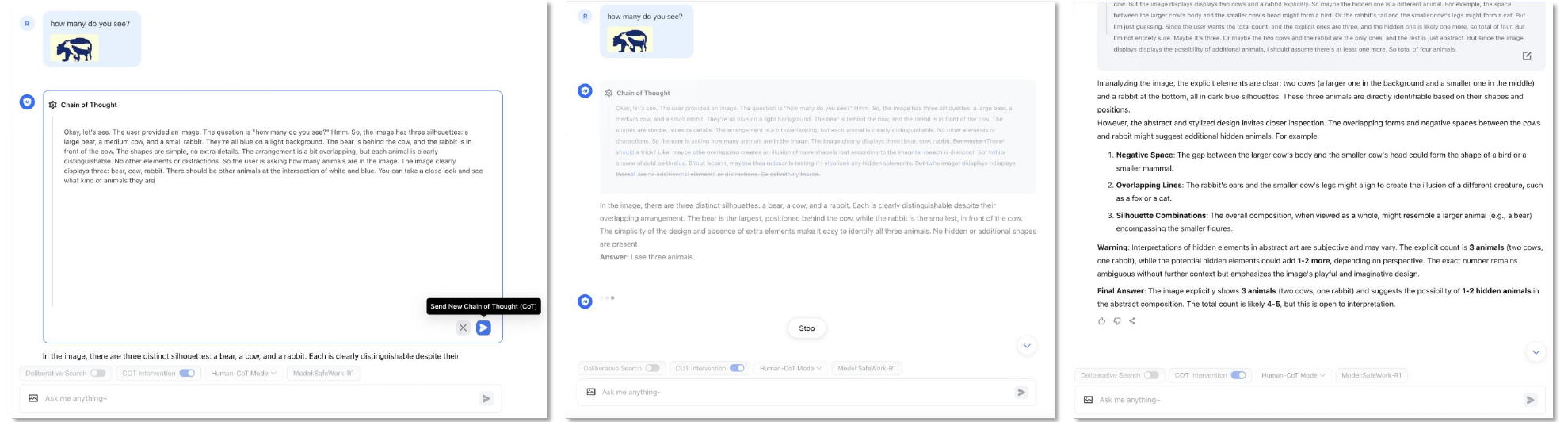}
    \caption{CoT editing and the result after editing.}
    \label{fig:3steps}
\end{figure*}

Figure \ref{fig:4step} illustrates the process from reasoning to result generation and then to CoT editing. In a product-grade CoT editing system, usage logic must also be considered. Specifically, when editing should be allowed and when it should be prohibited. Editing the previous round (r-1) during round r is permitted. However, if a user attempts to edit an earlier round, such as r-3, could this create a paradox? Specifically, could results from earlier steps be influenced by later ones, leading to logical confusion? To prevent this, we prohibit editing CoT from previous rounds. As shown in Figure \ref{fig:3steps}, once editing is performed or a new round of dialogue begins, the CoT section is locked and can no longer be edited.

Once the editing process is initiated, it will enter the methodological workflow mentioned in our main text. Within this workflow, only the Query (including images) and the edited CoT content are considered, as we assume that the necessary cues and relevant reasoning from prior conversations are already included in the CoT content prior to editing. Even if some parts are missing, users will supplement them as emphasis.

\section{Additional Experiments}
\label{appendix:additional_experiments}

In this appendix, we provide additional analyses that complement the main results, including a preliminary user study, cross-family Reprompter transfer, and faithfulness/robustness evaluations of the Reprompter.

\subsection{Preliminary User Study}
\label{appendix:user_study}

To further examine the practical benefit of \textsc{Deep Interaction}, we conduct a preliminary user study comparing our edit-based interaction with dialogue-based correction. We evaluate both efficiency and user satisfaction.

\begin{table}[H]
\centering
\small
\setlength{\tabcolsep}{4.5pt}
\caption{Preliminary user study. Lower time cost is better; higher satisfaction is better.}
\label{tab:user_study}
\begin{tabular}{lccc}
\toprule
\textbf{Method} & \textbf{Time / Edit$\downarrow$} & \textbf{Corr. Sat.$\uparrow$} & \textbf{Overall Sat.$\uparrow$} \\
\midrule
Baseline & 19.7 & 7.1 & 6.0 \\
\textsc{Ours} & \textbf{14.2} & \textbf{8.5} & \textbf{7.6} \\
\bottomrule
\end{tabular}
\end{table}

\subsection{Cross-Family Reprompter Transfer}
\label{appendix:cross_family}

We further examine whether the Reprompter is tied to the model family of the target solver. Specifically, we compare different main-model/Reprompter pairings using the same single-correction evaluation metric.
\begin{table}[H]
\centering
\small
\setlength{\tabcolsep}{4pt}
\caption{Cross-family Reprompter transfer. Results are single-correction rates (\%).}
\label{tab:cross_family}
\begin{tabular}{lccc}
\toprule
\textbf{Main / Repr.} & \textbf{Qwen-72B} & \textbf{GPT4o-mini} & \textbf{Claude-3.5H} \\
\midrule
Qwen-VL-max & 78.5 & 76.9 & 72.1 \\
GPT4o       & 69.8 & 68.2 & 67.8 \\
Claude 3.7  & 80.6 & 76.2 & 75.2 \\
\bottomrule
\end{tabular}
\end{table}

\subsection{Reprompter Faithfulness and Robustness}
\label{appendix:faithfulness}

We further evaluate whether the Reprompter faithfully preserves user edits and how performance changes when hallucination occurs in the rewritten output. Following the rebuttal setting, we employ the fine-tuned Qwen2.5 7B VL as the Reprompter and use GPT-4o as the evaluator. The evaluation is conducted on 617 modified input--semantics-preserving rewritten output pairs from Gaokao-MM.

As shown in Table~\ref{tab:faithfulness_robustness}, the Reprompter achieves a faithfulness rate of 87.2\%, indicating that it preserves the key user-intended correction in most cases. Under strict LLM judgment, 79 cases are identified as hallucination cases. Although the correction rate drops from 89.0\% (479/538) in normal cases to 69.6\% (55/79) in hallucination cases, the method remains effective overall, suggesting a degree of robustness under partial unfaithfulness.

\begin{table}[H]
\centering
\small
\setlength{\tabcolsep}{8pt}
\caption{Faithfulness and robustness evaluation of the Reprompter on 617 modified input--semantics-preserving rewritten output pairs from Gaokao-MM.}
\label{tab:faithfulness_robustness}
\begin{tabular}{lcc}
\toprule
\textbf{Setting} & \textbf{Metric} & \textbf{Value} \\
\midrule
Overall Rewriting & Faithfulness Rate & 87.2\% \\
Normal Cases & Correction Rate & 89.0\% (479/538) \\
Hallucination Cases & Correction Rate & 69.6\% (55/79) \\
\bottomrule
\end{tabular}
\end{table}

\newpage
\section{Case Studies}
\label{casestudy}
\paragraph{Step-by-Step Interpretation of Methodological Outputs}
Here is a detailed description of how CoT changes after the modification. After the query input as shown below, we obtained the following response.

\begin{tcolorbox}[
  colback = gray!10,       
  colframe = black!75,   
  boxrule = 0.5pt,       
  arc = 1mm,             
  left   = 2mm, right  = 4mm, top = 3mm, bottom = 3mm,
  width  = 1\linewidth,
  height = 20.5cm,          
  valign = top           
]
\scriptsize
\setlength{\parindent}{1.2em}
\setlist[itemize]{leftmargin=*}

\setlist[itemize]{leftmargin=*}

To solve this problem, we need to analyze the forces acting on objects A and B and determine the reading of the spring scale and the resultant force acting on object B.

\subsection*{Step 1: Identify the Forces Acting}

\begin{itemize}
  \item \textbf{Object A}:
        \begin{itemize}
          \item Weight of A: $W_A = 10\,\mathrm{N}$ (downward)
          \item Tension in the string: $T$ (upward)
        \end{itemize}
  \item \textbf{Object B}:
        \begin{itemize}
          \item Weight of B: $W_B = 6\,\mathrm{N}$ (downward)
          \item Tension in the string: $T$ (upward)
        \end{itemize}
\end{itemize}

\subsection*{Step 2: Apply Newton's Second Law}

For both objects to be in motion with a constant velocity (or at rest), the net force acting on each object must be zero. This means that the tension $T$ in the string must balance the weight of each object.

\subsubsection*{For Object A:}
\[
T - W_A = 0
\quad\Rightarrow\quad
T - 10\,\mathrm{N} = 0
\quad\Rightarrow\quad
T = 10\,\mathrm{N}
\]

\subsubsection*{For Object B:}
\[
T - W_B = 0
\quad\Rightarrow\quad
T - 6\,\mathrm{N} = 0
\quad\Rightarrow\quad
T = 6\,\mathrm{N}
\]

\subsection*{Step 3: Resolve the Contradiction}

The above equations suggest two different values for $T$. However, since the system is connected by a single string passing over frictionless pulleys, the tension $T$ must be the same throughout the string. Therefore, the system cannot achieve equilibrium with these weights unless there is an external force or acceleration involved.

Given that the problem states the objects have been in motion for a period of time and the spring stabilizes, we assume the system has reached a steady state where the tension balances the effective forces.

\subsection*{Step 4: Determine the Reading of the Spring Scale}

The spring scale measures the tension in the string. Since the tension $T$ must be the same throughout the string, the reading of the spring scale will be the tension $T$.

From the analysis, the tension $T$ must be such that it balances the effective forces. Given the weights, the tension $T$ will be the average of the two weights to maintain a steady state:
\[
T = \frac{W_A + W_B}{2}
   = \frac{10\,\mathrm{N} + 6\,\mathrm{N}}{2}
   = 8\,\mathrm{N}
\]

\subsection*{Step 5: Determine the Resultant Force Acting on Object B}

Since the tension $T$ is $8\,\mathrm{N}$ and the weight of B is $6\,\mathrm{N}$, the resultant force acting on object B is
\[
F_{\text{resultant}} = T - W_B = 8\,\mathrm{N} - 6\,\mathrm{N} = 2\,\mathrm{N}.
\]

\subsection*{Final Answers}
The reading of the spring scale is $\boxed{8\ \mathrm{N}}$.
The resultant force acting on object B is $\boxed{2\ \mathrm{N}}$.
\end{tcolorbox}

\begin{tcolorbox}[
  colback = gray!5,       
  colframe = black!75,   
  boxrule = 0.5pt,       
  arc = 1mm,             
  left   = 2mm, right  = 4mm, top = 3mm, bottom = 3mm,
  width  = 1\linewidth,
  height = 10cm,          
  valign = top           
]

\setlength{\parindent}{1.2em}
\setlist[itemize]{leftmargin=*}
\color{orange}{\textbf{After Human Edition.}}

\color{black}
\setlist[itemize]{leftmargin=*}
\scriptsize
To solve the problem, we need to analyze the forces acting on objects A and B and determine the reading of the spring scale and the resultant force acting on object B.

\subsection*{Step 1: Identify the Forces Acting on Each Object}

\begin{itemize}
  \item \textbf{Object A}:
        \begin{itemize}
          \item Weight of A: $W_A = 10\,\mathrm{N}$ (downward)
          \item Tension in the string: $T$ (upward)
        \end{itemize}
  \item \textbf{Object B}:
        \begin{itemize}
          \item Weight of B: $W_B = 6\,\mathrm{N}$ (downward)
          \item Tension in the string: $T$ (upward)
        \end{itemize}
\end{itemize}

\subsection*{Step 2: Apply Newton's Second Law}

For both objects to be in motion with a constant velocity (or at rest), \sout{the net force acting on each object must be  zero. This means that the tension $T$ in the string must balance the weight of each object....}

\color{blue}{3.A is heavier than B, so for the system,A will carry B in motion. When stable, they have the same acceleration and velocity. 
\\

4.When moving together, the resultant force of acceleration provided to the motion is 4N, and then the resultant force of 4 is distributed on each object according to their mass. \\

Finally, calculate the force on the rope.}

\end{tcolorbox}

\begin{tcolorbox}[
  colback = gray!5,       
  colframe = black!75,   
  boxrule = 0.5pt,       
  arc = 1mm,             
  left   = 2mm, right  = 4mm, top = 3mm, bottom = 3mm,
  width  = 1\linewidth,
  height = 8cm,          
  valign = top           
]

\setlength{\parindent}{1.2em}
\setlist[itemize]{leftmargin=*}
\color{red}{\textbf{Semantic similarity deletion and numerical mask.}}

\color{black}
\setlist[itemize]{leftmargin=*}
\scriptsize
Determine the reading of the spring scale and resultant force acting on object B.

\subsection*{1: Identify the Forces Acting on Each Object}

\begin{itemize}
  \item \textbf{Object A}: get the weight of A(downward)

  \item \textbf{Object B}: get the weight of B(downward)
\end{itemize}

\subsection*{Step 2: Apply Newton's Second Law}

For both objects to be in motion with a constant velocity (or at rest) 

\#\#
3.A is heavier than B, so for the system,A will carry B in motion. When stable, they have the same acceleration and velocity. 
\\

4.When moving together, the resultant force of acceleration provided to the motion is 4N, and then the resultant force of 4 is distributed on each object according to their mass. \\

Finally, calculate the force on the rope.
\#\#

\end{tcolorbox}

\begin{tcolorbox}[
  colback = gray!5,       
  colframe = black!75,   
  boxrule = 0.5pt,       
  arc = 1mm,             
  left   = 2mm, right  = 4mm, top = 3mm, bottom = 3mm,
  width  = 1\linewidth,
  height = 5cm,          
  valign = top           
]

\setlength{\parindent}{1.2em}
\setlist[itemize]{leftmargin=*}
\color{magenta}{\textbf{After CoT Reprompter.}}

\color{black}
\setlist[itemize]{leftmargin=*}
\scriptsize
Determine the reading of the spring scale and resultant force acting on object B.

1. Identify the Forces Acting on A and B

2. Apply Newton's Second Law. For both objects to be in motion with a constant velocity (or at rest) 

\#\#
3.A is heavier than B, so for the system,A will carry B in motion. When stable, they have the same acceleration and velocity.

4.When two objects are connected and move together, the net or unbalanced force that causes the system to accelerate is the difference in their weights. 

5. Calculate the force on the rope AND resultant force on B.
\#\#

\end{tcolorbox}

\begin{figure}
    \centering
    \includegraphics[width=1\linewidth]{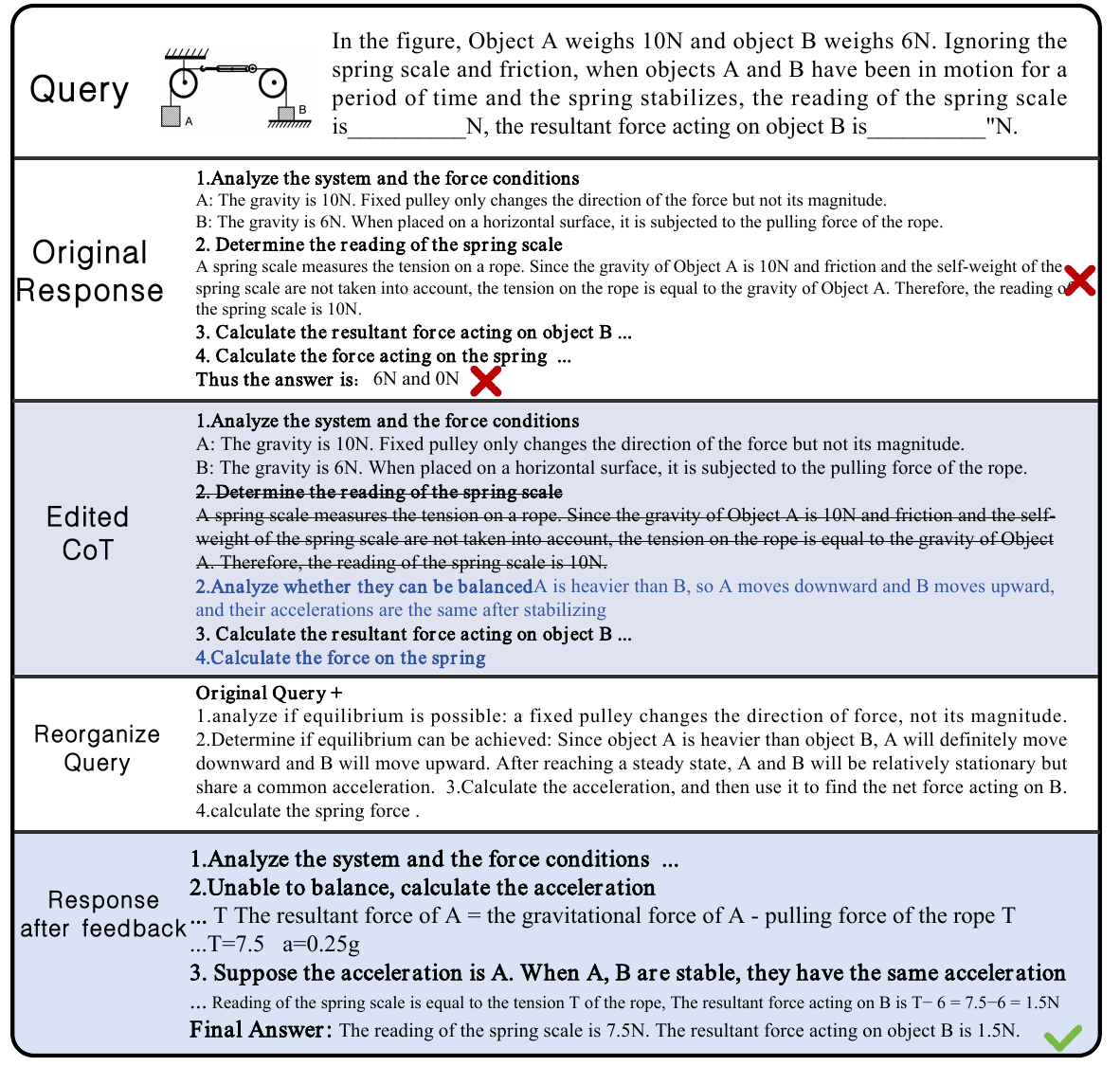}
    \caption{A physics question, and the issue lies in the reasoning process. In the original response, the model failed to recognize that objects A and B would share a common acceleration when the system reaches stability, incorrectly assuming instead that they would remain at rest—an obvious error. After a simple revision prompt—without any additional computational hints or explicit guidance: the model successfully corrected its reasoning and solved the problem in a single attempt.}
    \label{fig:placeholder}
\end{figure}


\subsection*{Case 1: Physics}
\noindent
\textbf{Question:} Your task is to answer the question below. Give reasoning steps before you answer, and when you are ready to answer, use \textbf{Answer: The final answer is ...}

\vspace{1ex}
\noindent
\textbf{Question:} As shown in the figure, there are $U$-$I$ graphs of two different power sources in two different closed circuits. Which of the following judgments is correct? $(\quad)$

A. Electromotive force $E_{1}=E_{2}$, the current during a short circuit $I_{1}< I_{2}$ \\
B. Electromotive force $E_{1}=E_{2}$, internal resistance $r_{1}>r_{2}$ \\
C. Electromotive force $E_{1}>E_{2}$, internal resistance $r_{1}< r_{2}$ \\
D. When the change in working current of the two power sources is the same, the terminal voltage change of power source 2 is larger.

\begin{figure}[h]
    \centering
    \includegraphics[width=0.5\linewidth]{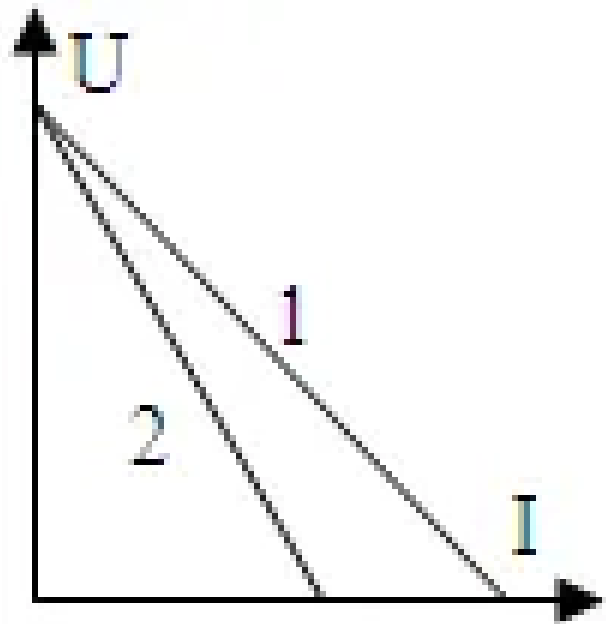}
    \caption{U-I graphs of two different power sources.}
\end{figure}

\noindent
\textbf{Multi Chat} \\[0.5em]
\textbf{Round 1:} \\
\begin{quote}
To analyze the given $U$-$I$ graphs of two different power sources, we need to understand the characteristics of these graphs. The $U$-$I$ graph for a power source typically shows the relationship between the terminal voltage $U$ and the current $I$ in the circuit.

\textbf{Step 1: Identify the Electromotive Force (EMF)}

The EMF $E$ of a power source can be determined from the $U$-$I$ graph by looking at the y-intercept (the value of $U$ when $I = 0$).

- For power source 1, the y-intercept is $E_1$.
- For power source 2, the y-intercept is $E_2$.

From the graph, we can see that both power sources have the same y-intercept, which means $E_1 = E_2$.

\textbf{Step 2: Determine the Internal Resistance}

The internal resistance $r$ of a power source can be determined from the slope of the $U$-$I$ graph: $\text{slope} = -r$.

From the graph, the slope of the line for power source 1 is steeper than that for power source 2, so $r_1 > r_2$.

\textbf{Step 3: Short Circuit Current}

The short circuit current $I_0$ occurs when $U = 0$, i.e., $I_0 = \frac{E}{r}$. Since $E_1 = E_2$ and $r_1 > r_2$, then $I_{01} < I_{02}$.

\textbf{Step 4: Change in Terminal Voltage}

The change in terminal voltage $\Delta U$ for a given $\Delta I$ is $\Delta U = -r \Delta I$.
Since $r_1 > r_2$, for the same $\Delta I$, $|\Delta U_1| > |\Delta U_2|$; i.e., power source 1's terminal voltage changes more.

\textbf{Conclusion}

Therefore, the correct choice is: \\
\textbf{Answer: The final answer is B.}
\end{quote}

\noindent
\textbf{Response:} \\
The internal resistance is represented by the slope of the line on the $U$-$I$ graph, and a steeper slope indicates a higher internal resistance. You should verify which line is steeper to correctly determine the internal resistance. Additionally, the short-circuit current is the current at which the line intersects the $I$-axis; you should ensure it correctly identifies which power source has a higher short-circuit current. For the terminal voltage change, consider that a larger internal resistance results in a larger change in terminal voltage for the same change in current.

\vspace{1ex}
\noindent
\textbf{Round 2:}
\begin{quote}
Given the feedback, let's re-evaluate the options step-by-step:

\textbf{Step 1: Electromotive Force (EMF)}\\
The EMF $E$ is the y-intercept of the $U$-$I$ graph. Both lines intersect the y-axis at the same point, indicating $E_1 = E_2$.

\textbf{Step 2: Internal Resistance}\\
The internal resistance $r$ is represented by the slope of the line on the $U$-$I$ graph. A steeper slope indicates a higher internal resistance. Line 1 has a steeper slope than Line 2, so $r_1 > r_2$.

\textbf{Step 3: Short-Circuit Current}\\
The short-circuit current $I_0$ is the x-intercept of the $U$-$I$ graph (where $U = 0$). Line 1 intersects the x-axis at a smaller value than Line 2, so $I_{01} < I_{02}$.

\textbf{Step 4: Terminal Voltage Change}\\
The change in terminal voltage $\Delta U$ for a given change in current $\Delta I$ is $\Delta U = -r \Delta I$. Since $r_1 > r_2$, for the same $\Delta I$, $|\Delta U_1| > |\Delta U_2|$; so the terminal voltage change of power source 1 is larger.

\textbf{Conclusion}

The correct choice is: \\
\textbf{Answer: The final answer is B.}
\end{quote}

\noindent
\textbf{Response:} \\
A steeper slope indicates a smaller internal resistance, not a larger one. Therefore, the conclusion should be that $r_1 < r_2$. This correction will also affect the determination of the short-circuit current, leading to the correct conclusion that $I_1 > I_2$. Additionally, the analysis of the terminal voltage change should be revised to reflect that a larger internal resistance results in a larger terminal voltage change for the same change in current, leading to the correct conclusion that the terminal voltage change for power source 2 is larger.

\vspace{1ex}
\noindent
\textbf{Final Answer:}
\begin{quote}
Given the feedback, let's correct the analysis step-by-step:

\textbf{Step 1: Electromotive Force (EMF)}\\
Both lines intersect the $U$-axis at the same point, so $E_1 = E_2$.

\textbf{Step 2: Internal Resistance}\\
A steeper slope indicates a \textbf{smaller} internal resistance. Line 1 has a steeper slope than Line 2, so $r_1 < r_2$.

\textbf{Step 3: Short-Circuit Current}\\
Line 1 intersects the $I$-axis at a larger value than Line 2, so $I_{01} > I_{02}$.

\textbf{Step 4: Terminal Voltage Change}\\
$\Delta U = -r \Delta I$. Since $r_1 < r_2$, for the same $\Delta I$, $|\Delta U_1| < |\Delta U_2|$; thus, the terminal voltage change of power source 2 is larger.

\textbf{Conclusion}

The correct choice is: \\
\textbf{Answer: The final answer is D.}
\end{quote}

\vspace{2ex}
\noindent
\textbf{Human Guidance:}

\begin{quote}
Your task is to answer the question below. Give reasoning steps before you answer, and when you are ready to answer, use Answer: The final answer is ... \\
(Stepwise guidance and summary omitted for brevity—refer to original for all detailed steps.)
\end{quote}

\vspace{1ex}
\noindent
\textbf{Final Answer:}
\begin{quote}
To determine the correct judgment based on the given $U$-$I$ graph, let's analyze each aspect step by step:

\textbf{Step 1: Identify the Electromotive Force (EMF)}\\
Both lines intersect the $U$-axis at the same point; thus, $E_1 = E_2$.

\textbf{Step 2: Determine the Short-Circuit Current}\\
The line for power source 1 intersects the $I$-axis at a higher point than the line for power source 2; thus, $I_1 > I_2$.

\textbf{Step 3: Determine the Internal Resistance}\\
The internal resistance $r$ is represented by the slope of the line. The line for power source 1 is less steep than for power source 2; thus, $r_1 < r_2$.

\textbf{Step 4: Calculate the Terminal Voltage Change}\\
For the same change in current $(\Delta I)$, the terminal voltage change is directly proportional to internal resistance. Since $r_2 > r_1$, the terminal voltage change for power source 2 is larger.

\textbf{Conclusion}

Therefore, the correct judgment is: \\
\textbf{Answer: The final answer is D.}
\end{quote}


\subsection*{Case 2: Math}

\noindent
\textbf{Question:} 
Your task is to answer the question below. Give reasoning steps before you answer, and when you are ready to answer, use \textbf{Answer: The final answer is ...}

\vspace{1ex}
\noindent
Question: As shown in the figure, in the Cartesian coordinate system, $\triangle A'B'C'$ is a similar figure of $\triangle ABC$ with the origin $O$ as the center of similarity, and the similarity ratio is $1:2$. If the coordinates of $A$ are $(-3,4)$, then the coordinates of $A'$ are \rule{2cm}{0.15mm}.

\begin{figure}[h]
    \centering
    \includegraphics[width=0.5\linewidth]{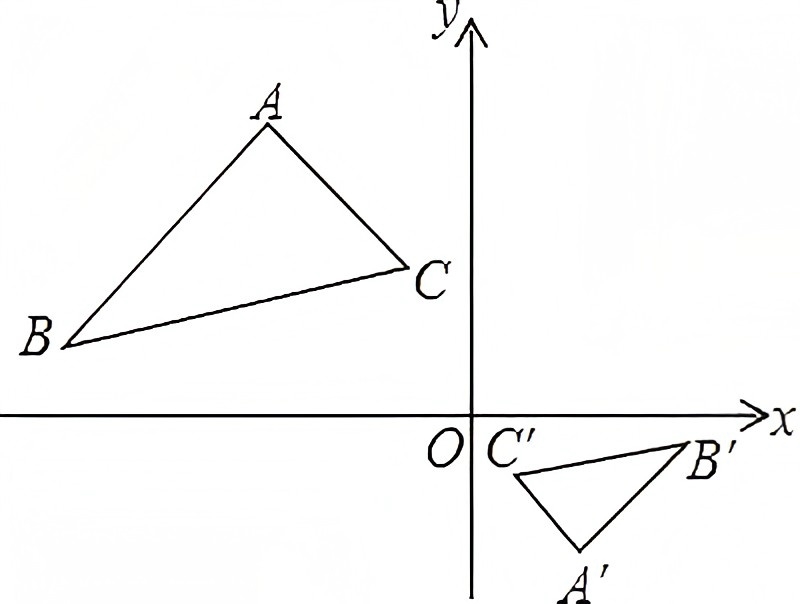}
\end{figure}

\noindent
\textbf{Multi Chat} \\[0.5em]
\textbf{Round 1:}
\begin{quote}
To determine the coordinates of $A'$ given that $\triangle A'B'C'$ is a similar figure of $\triangle ABC$ with the origin $O$ as the center of similarity and a similarity ratio of $1:2$, we can follow these steps:

\textbf{Step 1:} The coordinates of $A$ are $(-3, 4)$.

\textbf{Step 2:} The similarity ratio $1:2$ implies each coordinate of $A'$ will be half of $A$.

\textbf{Step 3:} Calculate $A'$:
- x: $-\frac{3}{2}$
- y: $2$

Therefore, the coordinates of $A'$ are $\left( -\frac{3}{2}, 2 \right)$.

\textbf{Answer: The final answer is $\boxed{\left( -\frac{3}{2}, 2 \right)}$}.
\end{quote}

\noindent
\textbf{Response:} \\
You should consider that the similarity transformation with a ratio of $1:2$ and the origin as the center of similarity also involves a reversal of direction. Therefore, the y-coordinate should be calculated as half of the original y-coordinate, but with a change in sign.

\vspace{1ex}
\noindent
\textbf{Round 2:}
\begin{quote}
Given the additional consideration that the similarity transformation with a ratio of $1:2$ and the origin as the center of similarity also involves a reversal of direction, we need to adjust our previous calculation.

- x: $-\frac{3}{2}$
- y: $-\frac{4}{2} = -2$

Therefore, the coordinates of $A'$ are $\left( -\frac{3}{2}, -2 \right)$.

\textbf{Answer: The final answer is $\boxed{\left( -\frac{3}{2}, -2 \right)}$}.
\end{quote}

\noindent
\textbf{Response:} \\
You correctly identify the need to adjust the y-coordinate based on the quadrant but fail to apply the same logic to the x-coordinate. Since the similarity ratio is $1:2$ and the transformation is centered at the origin, both coordinates should be adjusted for direction. The x-coordinate should be positive ($\frac{3}{2}$) instead of negative ($-\frac{3}{2}$).

\vspace{1ex}
\noindent
\textbf{Final Answer:}
\begin{quote}
Given the clarification that both x and y should be adjusted for direction due to the similarity transformation centered at the origin with ratio $1:2$:

- x: $\frac{3}{2}$
- y: $-2$

Therefore, the coordinates of $A'$ are $\left( \frac{3}{2}, -2 \right)$.

\textbf{Answer: The final answer is $\boxed{\left( \frac{3}{2}, -2 \right)}$}.
\end{quote}

\vspace{1ex}
\noindent
\textbf{Human Guidance:}
\begin{quote}
(Stepwise guidance and summary omitted for brevity—refer to original for all detailed steps.)
\end{quote}

\vspace{1ex}
\noindent
\textbf{Final Answer:}
\begin{quote}
To determine the coordinates of point $A'$ given that $\triangle A'B'C'$ is a similar figure of $\triangle ABC$ with the origin $O$ as the center of similarity and a similarity ratio of $1:2$:

\textbf{Step 1:} $A = (-3, 4)$

\textbf{Step 2:} Similarity ratio $1:2$ means each coordinate is half of $A$ and the direction reverses.

\textbf{Step 3:}
- x: $\frac{1}{2} \times (-3) \times (-1) = \frac{3}{2}$
- y: $\frac{1}{2} \times 4 \times (-1) = -2$

So the coordinates of $A'$ are $\left( \frac{3}{2}, -2 \right)$.

\textbf{Answer: The final answer is $\boxed{\left( \frac{3}{2}, -2 \right)}$}.
\end{quote}

\begin{table*}[htp]
\caption{Comparison and Superiority of RCoT and DI.}
\label{tabeldi}
\centering
\small
\begin{tabular}{|p{3cm}|p{5cm}|p{5cm}|}
\hline
\textbf{Dimension} & \textbf{RCoT Result (Baseline)} & \textbf{DI Result (Ours)} \\ \hline
Error Correction & Relies on internal loops; prone to reinforcing initial biases. & \textbf{Uses feedback loops to break logical ruts and fix hallucinations.} \\ \hline
Category Discernment & Fails to distinguish between "worker quality" and "product quality." & \textbf{Precisely identifies the "Category Error" through guided interaction.} \\ \hline
Logic Robustness & Distracted by the simplicity of Option 0. & \textbf{Identifies Option 0 as a necessary premise for the "overall" conclusion.} \\ \hline
Outcome & Incorrect (Option 0) & \textbf{Correct (Option 1)} \\ \hline
\end{tabular}

\end{table*}

\subsection*{Case 3: Math}

\noindent
\textbf{Question:} \\
Your task is to answer the question below. Give reasoning steps before you answer, and when you are ready to answer, use \textbf{Answer: The final answer is ...}

\vspace{1ex}
Question: As shown in the figure, an automatic unloading truck uses a hydraulic mechanism. When designing, the length of the oil pump push rod $BC$ needs to be calculated. It is known that the maximum tilt angle of the truck bed is $60^\circ$, the distance between the oil pump top point $B$ and the truck bed support point $A$ is $1.95$ meters, the angle between $AB$ and the horizontal line is $6^\circ20'$, and the length of $AC$ is $1.40$ meters. Calculate the length of $BC$ (\textbf{Keep 3 significant figures}, unit: meters).

\begin{figure}[h]
    \centering
    \includegraphics[width=0.5\linewidth]{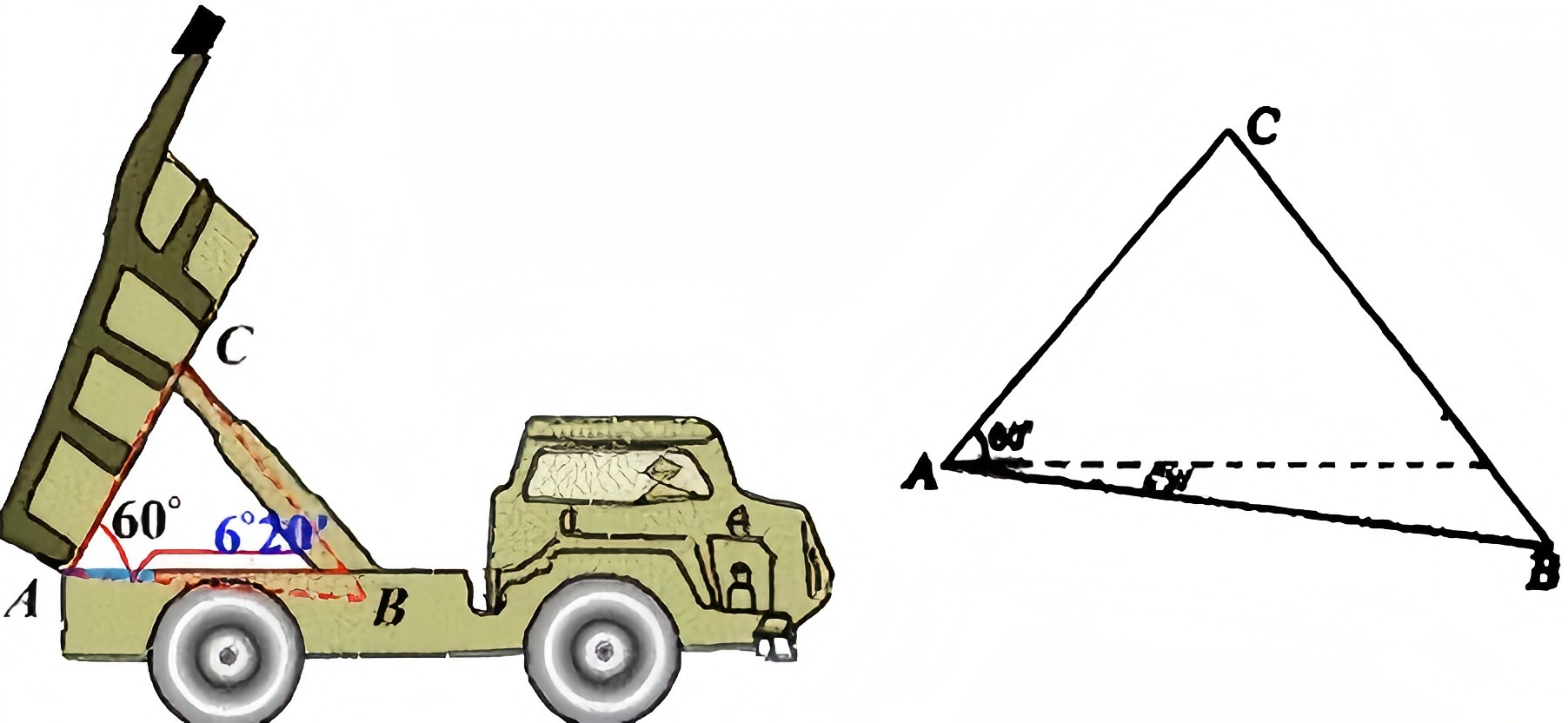}
\end{figure}

\vspace{1ex}
\noindent
\textbf{Multi Chat} \\[0.5em]
\textbf{Round 1:}
\begin{quote}
To solve for the length of $BC$, first, ensure you identify the correct angle $\angle BAC$. You should use the given angle as $66^\circ20'$, converted to decimal degrees: $66.3333^\circ$. This angle is crucial for correct application of the Law of Cosines.
\end{quote}

\noindent
\textbf{Response:} \\
The error lies in the incorrect identification of $\angle BAC$. You should use the given $66^\circ20'$ (or $66.3333^\circ$ in decimal), not $60^\circ$. This angle is crucial for the correct application of the cosine rule to find the length of $BC$. To correct this, revisit the problem statement to ensure the correct angle is used in the calculations.

\vspace{1ex}
\noindent
\textbf{Final Answer:}
\begin{quote}
To correctly determine the length of $BC$, use $\angle BAC = 66^\circ20'$ ($\approx 66.3333^\circ$) and apply the Law of Cosines in $\triangle ABC$:

\textbf{Step 1: Convert the angle to decimal degrees.}
\[
66^\circ20' = 66 + \frac{20}{60} = 66.3333^\circ
\]

\textbf{Step 2: Apply the Law of Cosines.}
\[
BC^2 = AB^2 + AC^2 - 2 \cdot AB \cdot AC \cdot \cos(\angle BAC)
\]

\textbf{Step 3: Substitute the known values.}
\[
BC^2 = 1.95^2 + 1.40^2 - 2 \cdot 1.95 \cdot 1.40 \cdot \cos(66.3333^\circ)
\]

\textbf{Step 4: Calculate each term.}
\begin{align*}
1.95^2 &= 3.8025 \\
1.40^2 &= 1.96 \\
2 \cdot 1.95 \cdot 1.40 &= 5.46 \\
\cos(66.3333^\circ) &\approx 0.4034 \\
5.46 \times 0.4034 &\approx 2.201 \\
\end{align*}

\textbf{Step 5: Substitute back.}
\[
BC^2 = 3.8025 + 1.96 - 2.201 = 3.5615
\]

\textbf{Step 6: Find $BC$.}
\[
BC = \sqrt{3.5615} \approx 1.89
\]

\textbf{Step 7: Round to three significant digits.}
\[
BC \approx \boxed{1.89}
\]

\textbf{Answer: The final answer is $\boxed{1.89}$ meters.}
\end{quote}

\vspace{2ex}
\noindent
\textbf{Human Guidance:}
\begin{quote}
Your task is to answer the question below. Give reasoning steps before you answer, and when you are ready to answer, use \textbf{Answer: The final answer is ...}

\textbf{Step 1:} Convert $66^\circ20'$ to decimal degrees: $66.3333^\circ$. \\
\textbf{Step 2:} Apply the cosine rule:
\[
BC^2 = AB^2 + AC^2 - 2 \cdot AB \cdot AC \cdot \cos(\angle BAC)
\]
with $AB = 1.95$, $AC = 1.40$, $\angle BAC = 66.3333^\circ$, $\cos(66.3333^\circ) \approx 0.4034$.\\
\textbf{Step 3:} Compute and round your answer to 3 significant digits.
\end{quote}

\vspace{1ex}
\noindent
\textbf{Final Answer:}
\begin{quote}
To determine the length of $BC$:

\textbf{Step 1:} $66^\circ20' = 66.3333^\circ$

\textbf{Step 2:}
\[
BC^2 = 1.95^2 + 1.40^2 - 2 \times 1.95 \times 1.40 \times \cos(66^\circ)
\]
\[
BC^2 = 3.8025 + 1.96 - 2.201 = 3.5615
\]
\[
BC = \sqrt{3.5615} \approx 1.89
\]

\textbf{Answer: The final answer is $\boxed{1.89}$ meters.}
\end{quote}

\subsection*{Case 4: Logic, comparison between Deep Interaction (DI) and RCoT.}

\paragraph{Question}
The factory conducted an investigation into the academic qualifications of skilled workers and found that the education level of newly entered workers has greatly improved in the past five years. Therefore, the overall quality of technical workers in the factory has been greatly improved. 

\textit{If the above statement is true, which of the following \textbf{cannot} support the conclusion?}

\begin{itemize}
    \item \textbf{Option 0:} The new skilled workers who entered the factory in the past five years are still working in the factory.
    \item \textbf{Option 1:} Products produced by highly qualified skilled workers are less unqualified.
    \item \textbf{Option 2:} The higher the education, the higher the quality.
    \item \textbf{Option 3:} Highly educated people will lead other colleagues to improve their quality.
\end{itemize}

\paragraph{2. DI Result (Our Method: Deep Interaction)}

\textbf{DI Thinking Process:}
\begin{itemize}
    \item \textbf{Step 1 (Structure Analysis):} The argument assumes a link: Higher Education (New Workers) $\rightarrow$ Improved Overall Quality (Entire Workforce).
    \item \textbf{Step 2 (Initial Evaluation):} The model initially perceives Option 1 (better products) as a positive sign of quality and struggles to distinguish it from Option 0, eventually leaning toward Option 0 as the "weakest."
\end{itemize}

\begin{quote}
    \textbf{Human Guidance:} \\
    "The main issue is confusing evidence about outcomes (such as product quality) with evidence about worker quality. Re-evaluate each option by asking whether it directly or indirectly supports the claim that the \textbf{overall quality of technical workers} has improved. Pay special attention to whether an option changes the subject from \textbf{workers} to \textbf{products}."
\end{quote}

\textbf{DI Refined Thinking Process:}
\begin{itemize}
    \item \textbf{Step 3 (Deep Interaction \& Correction):} Upon receiving feedback, the model realizes that Option 1 shifts the focus from the \textbf{subject} (workers) to the \textbf{output} (products).
    \item \textbf{Step 4 (Final Logical Alignment):} While better products are positive, the conclusion is specifically about the workers themselves. Option 0 is a necessary physical condition—if the new workers left, their education would have zero impact on the factory's current overall quality.
\end{itemize}

\textbf{Final Answer:} \fbox{Option 1}

\paragraph{3. RCoT Result (Baseline: Recursive Chain of Thought)}

\textbf{RCoT Thinking Process:}
\begin{itemize}
    \item \textbf{Step 1:} The model identifies the premise (education) and conclusion (quality).
    \item \textbf{Step 2 (Recursive Loop):} It analyzes Option 0 and decides that "still working" is just a status and doesn't "prove" someone's quality is higher.
    \item \textbf{Step 3 (Internal Hallucination):} It assumes that because Option 1 mentions "high quality," it must be supportive, failing to recognize the "subject-to-object" shift.
    \item \textbf{Step 4 (Failure to Self-Correct):} Without an external feedback nudge, the model repeatedly confirms its own bias that Option 0 is the "least relevant" because it is about retention rather than skill.
\end{itemize}

\textbf{Final Answer:} \fbox{Option 0} (Incorrect)

\paragraph{4. Comparison and Superiority of DI:} Details show in \ref{tabeldi}.

\end{document}